\documentclass[10pt,twocolumn,letterpaper]{article}

\usepackage[pagenumbers]{iccv} 

\usepackage{amsthm}
\usepackage{bbding}
\usepackage{pifont}
\usepackage{wasysym}
\usepackage{amssymb}
\usepackage{multirow}
\usepackage{graphicx}
\usepackage{colortbl}
\usepackage[ruled]{algorithm2e}
\usepackage{algorithmic}
\usepackage{adjustbox} 
\usepackage{float}
\usepackage{booktabs}
\usepackage{caption}
\usepackage{subcaption}

\definecolor{lightblue}{HTML}{3c78d8}
\definecolor{lightgreen}{HTML}{38761d}
\definecolor{lightpurple}{rgb}{0.9,0.85,1}

\newtheorem{definition}{Definition}

\definecolor{iccvblue}{rgb}{0.21,0.49,0.74}
\usepackage[pagebackref,breaklinks,colorlinks,allcolors=iccvblue]{hyperref}

\def\paperID{11426} 
\def\confName{ICCV}
\def\confYear{2025}

\title{CaO$_2$: Rectifying Inconsistencies in Diffusion-Based Dataset Distillation}

\author{%
  Haoxuan Wang$^{1}$\qquad
  Zhenghao Zhao$^{1}$\qquad
  Junyi Wu$^{1}$\qquad \\
  Yuzhang Shang$^{2}$\qquad
  Gaowen Liu$^{3}$\qquad
  Yan Yan$^{1}$\footnotemark[2]\\[4pt]
  \normalsize
  $^{1}$University of Illinois Chicago\quad
  $^{2}$University of Central Florida\quad
  $^{3}$Cisco Research\\[2pt]
}

\begin{document}
\maketitle
\begin{abstract}
The recent introduction of diffusion models in dataset distillation has shown promising potential in creating compact surrogate datasets for large, high-resolution target datasets, offering improved efficiency and performance over traditional bi-level/uni-level optimization methods. 
However, current diffusion-based dataset distillation approaches overlook the evaluation process and exhibit two critical inconsistencies in the distillation process: (1) Objective Inconsistency, where the distillation process diverges from the evaluation objective, and (2) Condition Inconsistency, leading to mismatches between generated images and their corresponding conditions. 
To resolve these issues, we introduce \textbf{C}ondition-\textbf{a}ware \textbf{O}ptimization with \textbf{O}bjective-guided Sampling (\textbf{CaO$_2$}), a two-stage diffusion-based framework that aligns the distillation process with the evaluation objective. 
The first stage employs a probability-informed sample selection pipeline, while the second stage refines the corresponding latent representations to improve conditional likelihood.
CaO$_2$ achieves state-of-the-art performance on ImageNet and its subsets, surpassing the best-performing baselines by an average of 2.3\% accuracy. 
Code is available at \url{https://github.com/hatchetProject/CaO2}.
\end{abstract}

\renewcommand{\thefootnote}{\fnsymbol{footnote}}
\footnotetext[2]{Corresponding author}

\section{Introduction}
\label{sec:intro}
The rapid expansion of data scale has significantly advanced the development of machine learning, but has also placed substantial demands on the storage capacity and computational resources. To accelerate training and reduce storage requirements while maintaining comparable performance, dataset distillation (DD) \cite{dd,dd_survey,mtt,ltdd,dq} was introduced to construct a compact surrogate dataset that captures the most critical information from a large target dataset. Conventionally, dataset distillation was formulated as a bi-level/uni-level optimization problem, designed to match the training dynamics between a teacher model trained on the target dataset and a student model trained on the synthetic dataset \cite{dream,cafe,datm,sre2l}. However, these matching-based methods either struggle to scale up to larger, higher-resolution datasets, or suffer from low performance.

\begin{figure}[t]
\begin{center}
\centerline{\includegraphics[trim=95 15 105 30,clip, width=1.0\linewidth]{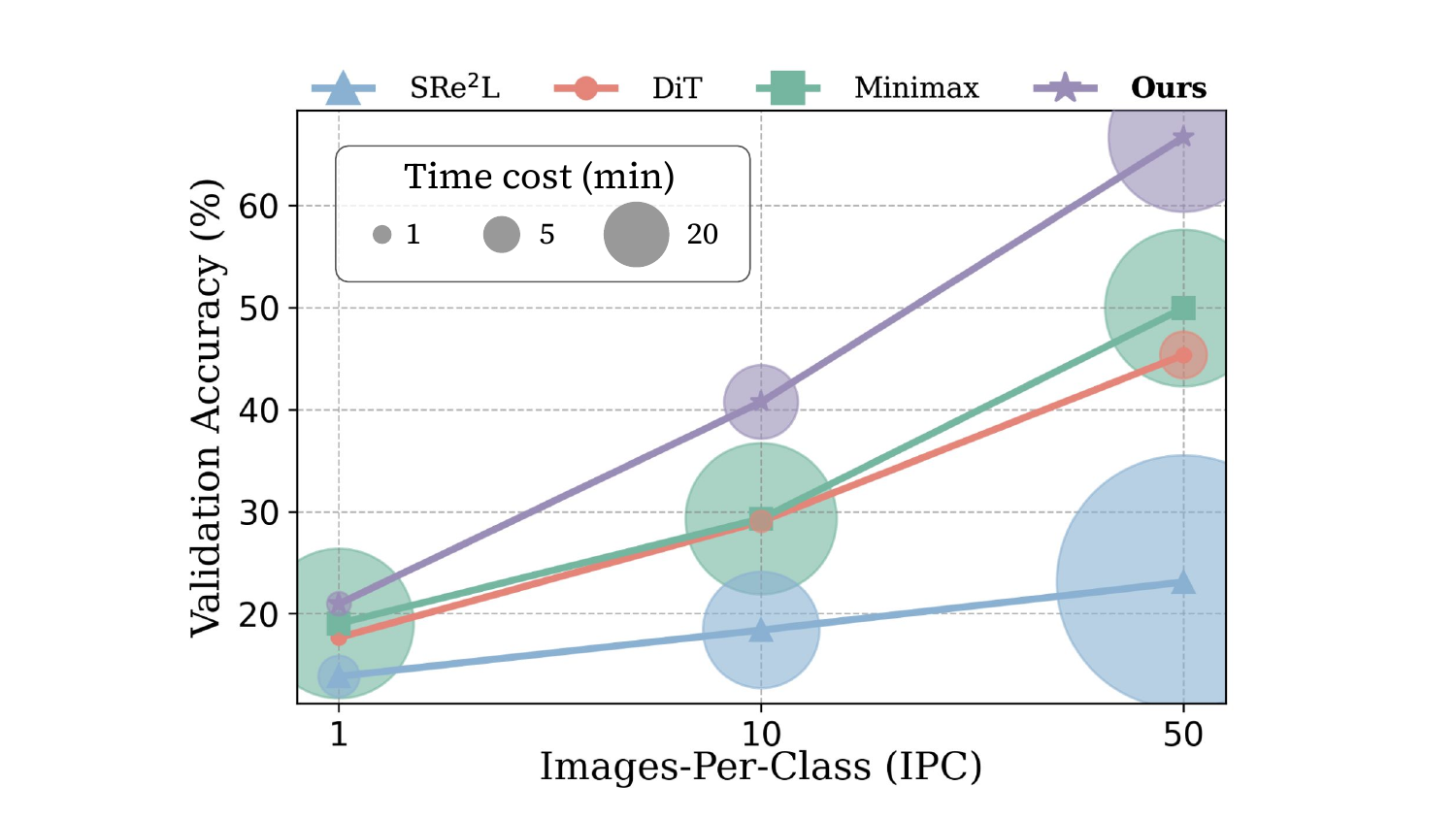}}
\vspace{-8pt}
\caption{\textbf{Comparison of validation accuracy and distillation time for different methods under different IPCs on ImageWoof.} Our two-stage method is more efficient and can obtain better performance compared with other SOTA approaches.}
\label{fig:time_acc}
\end{center}
\vspace{-32pt}
\end{figure}

Recently, diffusion-based dataset distillation \cite{minimax,d4m} emerged as a new paradigm for efficient data condensation. These methods utilize pre-trained diffusion models \cite{dit,sd} as strong distribution learners to filter noisy representations and retain the most important ones \cite{glad}. Therefore, their generated images are strong representations of the original target data distribution. As shown in \cref{fig:time_acc}, directly using randomly sampled images from the Diffusion Transformer (DiT) \cite{dit} for evaluation already outperforms the uni-level optimization method SRe$^2$L \cite{sre2l} by a significant margin, both in performance and distillation time. Moreover, unlike matching-based methods that rely on an additional student model and output unreadable images, diffusion-based methods are independent of the student model and produce realistic images. This enables the synthetic dataset to be seamlessly used for other tasks, such as neural architecture search \cite{darts} and continual learning \cite{continual}.

These compelling advantages, nevertheless, come at the cost of achieving optimal performance. We observe that current diffusion-based DD methods \cite{minimax,d4m} overlook the evaluation objective that assesses the discriminative properties of the distilled images. Instead, they focus exclusively on enhancing the fidelity of generated images, either through diffusion model training \cite{minimax} or latent embedding clustering \cite{d4m} (as shown in \cref{fig:pipeline}(a)). Such an oversight may lead to inappropriately designed distillation objectives, ultimately resulting in suboptimal evaluation accuracy. 

However, incorporating the evaluation objective into the distillation process is not trivial, as diffusion models are not designed for classification tasks. Rather than directly integrating the evaluation objective, we aim to address two key inconsistencies related to it. 
The first is Objective Inconsistency, referring to the misalignment between the objective of the generative diffusion model and that of the evaluation model. As shown in \cref{fig:pipeline}(a), matching-based methods \cite{dc,idc,mtt} constrain the distilled dataset to be optimized with the same objective of image classification as in the evaluation phase. In contrast, current diffusion-based methods generate the distilled dataset without any classification supervision.
The second is Condition Inconsistency, stemming from the limitations of the applied conditional diffusion models themselves. In practice, diffusion models are not perfectly trained where the conditional likelihood for every generated sample is not maximized. Consequently, the obtained image-label pairs are suboptimal, which adversely impacts the training process during evaluation.

Having revealed the inconsistencies, we alleviate them by proposing a two-stage framework named \textbf{C}ondition-\textbf{a}ware \textbf{O}ptimization with \textbf{O}bjective-guided Sampling (\textbf{CaO$_2$}). As illustrated in \cref{fig:pipeline}(b), the first stage generates an image pool and sample from it class-wisely. By using a pre-trained lightweight classifier, we ensure that only samples confidently classified as belonging to their conditioned class are selected, thereby easing Objective Inconsistency. In the second stage, the chosen image latent is perturbed with random noise and optimized while keeping the diffusion model fixed under a task-dependent condition. The optimization objective aims to maximize the conditional likelihood of the image with respect to its condition, thereby reducing Condition Inconsistency. 
Our approach achieves state-of-the-art performance with better efficiency on ImageNet and its subsets. We also show that apart from pre-trained diffusion models, our method can be applied to finetuned models \cite{minimax} and autoregressive models \cite{mar}.

In summary, we make the following contributions:
\begin{itemize}
    \item We observe that current diffusion-based dataset distillation methods overlook the evaluation process, leading to two inconsistencies in the distillation process — Objective Inconsistency and Condition Inconsistency — that hinder effective data condensation.
    \item To mitigate the inconsistencies, we propose CaO$_2$, a two-stage framework enabling efficient distillation with improved performance. Our framework can be applied to different model backbones, including Diffusion Transformers and autoregressive generation models such as MAR. It can also be used as a plug-and-play module for existing diffusion-based dataset distillation methods.
    \item Empirical results indicate that CaO$_2$ achieves state-of-the-art performance on ImageNet and its subsets, outperforming other methods by an average of 2.3\% accuracy. 
\end{itemize}
\section{Related Works}
\label{sec:related}

\subsection{Matching-based Dataset Distillation}
Initial works have formulated the dataset distillation problem as a bi-level optimization task, aiming to match the training characteristics of the synthetic dataset with those of the target dataset, such as gradients \cite{dc,idc,dream}, feature distributions \cite{cafe,dm,idm}, and training trajectories \cite{mtt,tesla,ftd,datm}. However, the bi-level distillation objective introduces large computational costs, making it impractical to generalize to large-scale datasets such as ImageNet. Even for a small dataset such as CIFAR-100, adopting the efficient matching-based method TESLA \cite{tesla} to distill a subset with IPC=10 can take nearly 20 hours.

To scale efficiency, SRe$^2$L \cite{sre2l} and its subsequent works \cite{cda,scdd,gvbsm,edc} adopted a uni-level optimization framework. They first synthesize images by matching model outputs and Batch Norm statistics using a well-trained teacher model, then generate multiple soft labels for the distilled images. Though efficiency is improved, their performance is limited due to the inadequate matching objective and the separate processes for image and label synthesis.

\begin{figure*}[t]
\begin{center}
\centerline{\includegraphics[trim=60 35 55 20,clip, width=1.0\textwidth]{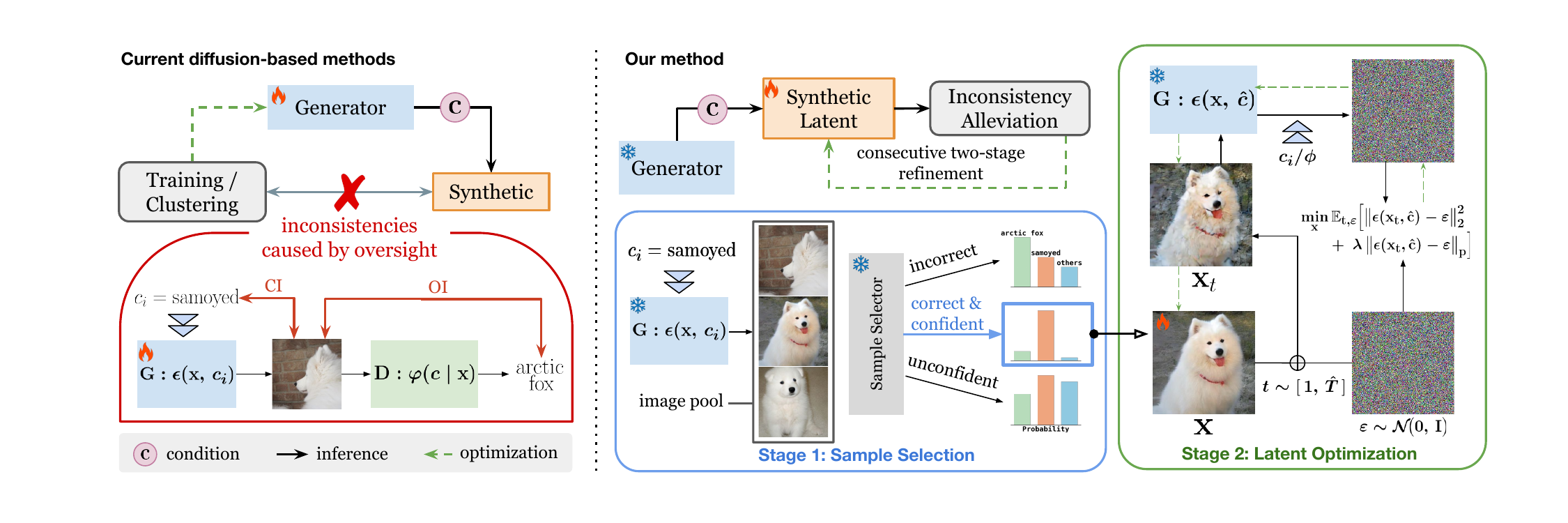}}
\hspace{-2.2cm} \textbf{(a)} Previous diffusion-based methods
\hspace{2.5cm} \textbf{(b)} Illustration of our two-stage framework
\vspace{-0.1in}
\caption{\textbf{Comparison and illustration of previous diffusion-based methods and our approach.} 
\textbf{(a)} Current diffusion-based methods facilitate efficient distillation using a conditional diffusion model $\mathbf{G}$. However, their proposed distillation processes overlook the evaluation objective, resulting in Objective Inconsistency (OI) between distillation and evaluation, as well as Condition Inconsistency (CI)  between the input condition and the generated image.
\textbf{(b)} To alleviate the inconsistencies, we propose a two-stage framework: Stage 1 mitigates OI by generating an image pool and selecting more discriminative samples through a lightweight sample selector. Stage 2 refines the selected image latents using \cref{eq:objective} and \cref{eq:condition}, reducing CI by maximizing the conditional likelihood of the generated samples. By avoiding training of the generative backbone, our method is efficient and applicable across different generative models.}
\label{fig:pipeline}
\end{center}
\vspace{-0.3in}
\end{figure*}

\subsection{Diffusion-based Dataset Distillation}
An emerging line of research \cite{minimax,d4m,igd} utilizes pre-trained generative models for dataset distillation. GLaD \cite{glad}, as a pioneering work, argues that images distilled into the latent space are less noisy and have clearer visual structures than the images in pixel space, and uses generative models for producing deep priors to integrate with matching-based methods. Minimax Diffusion \cite{minimax} instead fully utilizes pre-trained diffusion models for generating realistic synthetic datasets. It finetunes the diffusion model with a minimax criterion to enhance sample representativeness and diversity. D$^4$M \cite{d4m} uses pre-trained text-to-image diffusion models and adopts prototype learning to learn cluster centers for each category. These diffusion-based dataset distillation methods achieved the new state-of-the-art performance on large-scale datasets such as ImageNet, significantly outperforming the current matching-based methods in both efficiency and performance. 

However, the distillation process in current diffusion-based dataset distillation methods is independent of the evaluation objective, with limited attention to the constraints posed by the small number of samples in the synthetic dataset. To address this discrepancy, our method introduces a two-stage framework incorporating sample selection and latent optimization, designed to achieve a more precise and consistent distillation process.
\section{Methodology}
\label{sec:method}
\subsection{Preliminary}
For a conditional diffusion model with parameter $\theta$, the process of generating a sample $\mathbf{x}_{0}$ adheres to a specific Markov chain structure:
\begin{equation}
    p_{\theta}(\mathbf{x}_0 | \mathbf{c}) = \int_{\mathbf{x}_{1:T}} p(\mathbf{x}_T) \prod_{t=1}^{T} p_{\theta}(\mathbf{x}_{t-1} | \mathbf{x}_t, \mathbf{c}) \, \mathrm{d}\mathbf{x}_{1:T},
    \label{eq:markov}
\end{equation}
where $\mathbf{c}$ is the input condition, $T$ is the number of denoising steps, and $p(\mathbf{x}_T)$ is the standard normal distribution. Directly maximizing the conditional likelihood $p_{\theta}(\mathbf{x}_0 | \mathbf{c})$ is impractical, thus diffusion models instead learn to optimize the variational lower bound (ELBO) on log-likelihood:
\begin{equation}
    \log p_{\theta}(\mathbf{x}_0 | \mathbf{c}) \geq - \mathbb{E}_{t, \varepsilon} [ \left\| \varepsilon - \epsilon_\theta \left( \mathbf{x}_t, \mathbf{c}, t \right) \right\|^2 ].
    \label{eq:elbo}
\end{equation}
Given the objective of likelihood maximization, an optimal diffusion model is expected to provide accurate density estimation for each class over the entire training data distribution. This implies that diffusion models are inherently strong classifiers \cite{diffusion_classifier,robust_classification}, and can efficiently generate samples that represent the underlying data distribution well.

However, the effectiveness of current diffusion-based methods is limited by the misalignment between the distillation and evaluation process. In \cref{sec:ss} and \ref{sec:lo}, we analyze two types of inconsistencies and address each to improve distillation performance. Motivated by bi-level optimization approaches, we introduce task-oriented variations in \cref{sec:task} for better method adaptation. Finally, in \cref{sec:mar}, we demonstrate how our method can be extended to utilize the masked autoregressive generation model.

\subsection{Objective-guided Sample Selection}
\label{sec:ss}
A conditional diffusion model takes a class condition $c_{i}$ as input to generate realistic images $\mathbf{x}_0$ from sampled noises. Assume the conditional diffusion model is trained perfectly with accurate density estimations so that $p_{\theta}(\mathbf{x}_0 | c_{i})$ is maximized. Then by applying Bayes' theorem as $p_{\theta}(c_{i}|\mathbf{x}_0) \propto p_{\theta}(\mathbf{x}_0 | c_{i}) p(c_{i})$, we can conclude that the conditional probability $p_{\theta}(c_{i}|\mathbf{x}_0)$ is also maximized. However, this equivalence in likelihood maximization holds only for diffusion classifiers \cite{diffusion_classifier} with the objective in \cref{eq:elbo}, and there is no guarantee that for a typical discriminative model $\varphi$ with a classification objective, $p_{\varphi}(c_{i}|\mathbf{x}_0)$ can be maximized concurrently. Furthermore, \cite{diffusion_classifier} empirically demonstrated that, even when the discriminative model is trained on a synthetic dataset of equivalent size to the full training set, the diffusion classifier still surpasses the discriminative model by 4.8\% in accuracy on ImageNet.

Moreover, randomness in noise initialization and sampling can produce low-quality, unrepresentative images, occasionally generating samples that are unhelpful for training or even poisonous. This issue is especially pronounced in low IPC settings, where each sample constitutes a significant portion of the dataset and is crucial for effective training.
Therefore, we reveal the \textbf{Objective Inconsistency (OI)} in diffusion-based dataset distillation, defined as: 
\begin{definition}
Let $c_{i}$ denote the input condition, $\{\mathbf{x}_{0}^{i,k}\}_{k=1}^{N}$ represent the set of $N$ images generated according to condition $c_{i}$, and $\{c^{k}\}_{k=1}^{N}$ denote the corresponding set of hard labels predicted by a trained classifier. Objective Inconsistency arises when there are mismatches between the input condition and the output label, i.e., $\exists k$, such that $c_i \neq c^{k}$.
\end{definition}
The Objective Inconsistency illustrates the discrepancy between the objectives of the distillation and evaluation stages in diffusion-based dataset distillation. It implies that some generated images, when used for classification, will be classified to belong to a different class than their conditioned label, impacting the correctness and fidelity of the distilled dataset. To alleviate this issue, we introduce a simple yet effective \textbf{sample selection} strategy based on classification probability to improve distillation performance. As shown in \cref{fig:pipeline}(b), while conventional methods directly generate the compact dataset, we adopt a post hoc approach by introducing a lightweight classifier to determine better samples for learning with the classification objective.

Specifically, to distill samples for a single class with a given IPC, we first generate an image pool of size $m \times \text{IPC}$ using a pre-trained diffusion model conditioned on that class, where $m$ is a scaling factor. Next, we use a lightweight pre-trained classifier (e.g., ResNet-18) to obtain predictive probabilities for each generated image, selecting only those samples predicted to belong to the conditioned class. Building on insights from \cite{dd_sample_diff}, which suggests prioritizing easier samples under lower IPC settings and harder samples for higher IPC settings, we further refine our selection by choosing the top-IPC most or least confident samples from the correct predictions. For classes lacking sufficient correct samples, we supplement the distilled dataset by randomly selecting from the remaining images. Additionally, we observe that using small scaling factors, such as $m=2$ or $m=4$, typically suffices to achieve strong performance, underscoring the efficiency of our strategy.

\subsection{Condition-aware Latent Optimization}
\label{sec:lo}
A good compact dataset expects each training image accurately reflecting its corresponding label. However, empirical diffusion models often fail to provide fully accurate density estimates of $p_{\theta}(\mathbf{x}_0 | \mathbf{c})$ for all conditions, primarily due to the errors in approximating the ELBO and the presence of the non-zero diffusion training loss. Consequently, even after sample selection, the generated samples remain sub-optimal for conditional likelihood maximization when using hard labels as input conditions. We define this inherent discrepancy between the input condition and the output likelihood as the \textbf{Condition Inconsistency (CI)}:
\begin{definition}
Let $\mathbf{x}_{0}^{i}$ represent the generated image latent conditioned on class $c_i$, $p_{\theta}(\mathbf{x}_0|\mathbf{c})$ be the conditional likelihood of the empirical diffusion model.
Condition Inconsistency indicates that $\forall i, \exists j \neq i$ such that $p_{\theta}(\mathbf{x}_{0}^{i}|c_{j}) > 0$.
\end{definition}

This definition implies that the image latent generated by an empirical diffusion model is not exclusively associated with its specified condition, thereby weakening the correlation between the sample and its label in the distilled dataset. Since a diffusion model with a lower diffusion loss approximates the conditional likelihood more accurately, one straightforward approach to alleviate Condition Inconsistency is to learn a well-trained diffusion model for each class. However, this solution is computationally prohibitive and impractical given the limited amount of samples and the large number of existing conditions.

Instead of optimizing the model itself, we propose to update the generated images through \textbf{latent optimization}. By minimizing the diffusion loss with respect to the input $\mathbf{x}$, we allow the input to shift toward regions where the pre-trained diffusion model yields more accurate density estimates. Therefore, given the input Gaussian noise $\varepsilon \sim \mathcal{N}(\mathbf{0}, \mathbf{I})$, input condition $\mathbf{\hat{c}}$ and the noisy input at time step $t$, the optimization objective is formulated as:
\begin{equation}
    \min_{\mathbf{x}} \mathbb{E}_{t,\varepsilon} \left[ \| \epsilon_{\theta}(\mathbf{x}_t, \mathbf{\hat{c}}, t) - \varepsilon \|_2^2 \right], \text{s.t. } \mathbf{x}_{t}=\sqrt{\overline{\alpha}_t}\mathbf{x} + \sqrt{1-\overline{\alpha}_t}\varepsilon.
    \label{eq:obj_wo_reg}
\end{equation}
$\overline{\alpha}_t$ is a constant with respect to $t$, and $t$ is randomly sampled from [1, $\hat{T}$], where $\hat{T} \ll T$. This suggests that the latent is only moderately perturbed. However, solely using the above objective may lead to optimized samples $\overline{\mathbf{x}}$ falling into regions associated with other classes \cite{robust_classification}. To constrain the degree of update, we use the max norm as additional regularization such that $\| \mathbf{\overline{\mathbf{x}}} - \mathbf{x} \|_\infty \leq \eta$. Thus, the final optimization objective is:
\begin{equation}
    \min_{\mathbf{x}} \mathbb{E}_{t,\varepsilon} \left[ \| \epsilon_{\theta}(\mathbf{x}_t, \mathbf{\hat{c}}, t) - \varepsilon \|_2^2 + \lambda \| \epsilon_{\theta}(\mathbf{x}_t, \mathbf{\hat{c}}, t) - \varepsilon \|_\infty \right],
    \label{eq:objective}
\end{equation}
where $\lambda$ controls the strength of regularization.
Note that the above is conducted in the latent space and, for the sake of simplicity, we refrain from introducing new notations to distinguish between the image and its latent.

\subsection{Task-oriented Variation}
\label{sec:task}
When computationally feasible, bi-level optimization methods exhibit strong performance \cite{datm,ncfm}, primarily because aligning the learning objectives during distillation and evaluation naturally removes the need to redesign learning strategies.
In contrast, non-matching-based methods enforce a strict separation between these two processes, preventing distillation-stage settings from benefiting the evaluation process. This discrepancy motivates us to refine the proposed two-stage approach with task-oriented designs.

Inspired by \cite{curriculum}, we use the size of the distilled dataset as an evidence during the sample selection stage. Since training benefits from progressively learning more difficult samples, we propose emphasizing easier samples when training on more challenging tasks, while prioritizing harder samples for simpler tasks. Consequently, we select correct and highly confident samples in lower IPC settings, and correct but less confident samples in higher IPC settings. 

\begin{algorithm}
\caption{Pseudocode for CaO$_2$}
\label{alg:loss}
\textbf{Input:} Pre-trained diffusion model $\epsilon(\mathbf{x}, \mathbf{c})$, sample selector $s(\mathbf{c}|\mathbf{x})$, target category set $\mathbb{C}$, IPC=$N$ \\
\textbf{Initialize:} Distilled dataset $\mathcal{S}=\{\}$, scaling factor $m$, optimization condition $\hat{\mathbf{c}}$\\
    \nl \For{$c \in \mathbb{C}$}{
        \tcp{\textcolor{lightblue}{Sample Selection}}
        \nl Obtain image pool $\mathcal{X}$ of size $mN$ from $\epsilon(\mathbf{x}, c, t)$ \\
        \nl Calculate probabilities $\mathcal{P}=\text{softmax}(s(\mathbf{c}|\mathcal{X}))$ \\
        \nl Select the top-$N$ correct and least/most confident samples $\mathcal{X}_{N}$ from $\mathcal{X}$ using $\mathcal{P}$ \\
        \tcp{\textcolor{lightgreen}{Latent Optimization}}
        \nl Sample a random Gaussian noise $\varepsilon \sim \mathcal{N}(0,\mathbf{I})$ \\
        \nl \For{$\mathbf{x} \in \mathcal{X}_{N}$}{
            \nl \For{each iteration}{
                \nl Sample a random time step $t$  \\
                \nl Obtain $\mathbf{x}_{t} = \sqrt{\overline{\alpha}_t}\mathbf{x} + \sqrt{1-\overline{\alpha}_t}\varepsilon$ \\
                \nl Update $\mathbf{x}$ using \cref{eq:objective}, \cref{eq:condition}
            }
            \nl $\mathcal{S} = \mathcal{S} \cup \{(\mathbf{x}, c)\}$ 
        }
    }
\textbf{Output:} Distilled dataset $\mathcal{S}$
\end{algorithm}

In the latent optimization stage, we examine the varying modeling quality across categories in diffusion models, and propose choosing different values for the input condition $\mathbf{\hat{c}}$. This approach is motivated by the fact that $\mathbf{\hat{c}}$ carries class-specific information that is inherently discriminative. For the set of classes $\mathbb{C}_{e}$ in an easier task and the set of classes $\mathbb{C}_{h}$ in a harder task, different values are considered: 
\begin{equation}
    \mathbf{\hat{c}} = c \cdot 1(c \in \mathbb{C}_{e}) + \phi \cdot 1(c \in \mathbb{C}_{h})
    \label{eq:condition}
\end{equation}
where $c$ is the true class label used for generating the latent, $\phi$ is the unconditional label used for classifier-free guidance and $1(\cdot)$ is the indicator function. For relatively easier tasks, we assume the input conditions can effectively provide discriminative information, and thus the true labels are used for guidance during optimization. For harder tasks, the guidance provided by the conditions may be insufficient or even detrimental, and thus classifier-free guidance is adopted. In this case, $\epsilon_\theta$ is treated as a well-trained single-step denoiser. We utilize it so that, even without conditional guidance, the latent is optimized to approximate the result achieved with conditional guidance, implying that conditional information is embedded within the image latent. 


During implementation, we examine the validation accuracy of each classification task to determine the task difficulty and the specific strategy to be used. The complete procedure of our algorithm is depicted in \cref{alg:loss}. 

\subsection{Extending the Generation Backbone}
\label{sec:mar}
Apart from conventional diffusion models, the Masked Autoregressive model (MAR) \cite{mar} has recently been proposed to reinvent autoregressive image generation. MAR utilizes an autoregressive procedure to predict image patches, with a diffusion loss guiding the training process. Motivated by the similarity in the learning objective, we show that apart from conventional diffusion models, our method can also be seamlessly applied to MAR for effective dataset distillation. 

While the sample selection stage remains similar, there are two major differences in the latent optimization stage compared with the diffusion model backbone. One is that instead of adding random Gaussian noises to the image latent with respect to the time step, we apply random masks on the latents with a maximum token masking ratio. The other difference is that MAR did not design a specific embedding for the classifier-free guidance, thus we replace $\phi$ with a self-designed zero label embedding for \cref{eq:condition}. The optimization objective can thus be formulated as:
\begin{equation}
    \min_{\mathbf{x}} \mathbb{E}_{\mathbf{m},\varepsilon} \left[ \| \epsilon_{\theta}(\mathbf{x}_{\mathbf{m}}, \mathbf{\hat{c}}) - \varepsilon \|_2^2 + \lambda \| \epsilon_{\theta}(\mathbf{x}_{\mathbf{m}}, \mathbf{\hat{c}}) - \varepsilon \|_p \right],
    \label{eq:mar_objective}
\end{equation}
where $\mathbf{m} \sim \mathcal{M}(0, \mathbf{R})$ is a randomly sampled mask with a maximum masking ratio of $\mathbf{R}$, $\mathbf{x}_{\mathbf{m}}$ is obtained by randomly masking $|
\mathbf{m}|$ tokens from $\mathbf{x}$. The workflow of our MAR-based approach is illustrated in Appendix \ref{sec:mar_supplement}. 

\section{Experiments}
\label{sec:experiment}
In this section, we clarify the experiment settings and the implementation details, perform comparisons with state-of-the-art methods and diffusion-based methods, and provide extensive ablation studies for discussion and analysis.

\begin{table*}[t]
\centering
\small
\adjustbox{width=1.0\textwidth}{
\begin{tabular}{cc|cccc|cccc}
\toprule
\multirow{2}{*}{Settings}   & \multirow{2}{*}{IPC} & \multicolumn{4}{c|}{ImageWoof} & \multicolumn{4}{c}{ImageNette} \\
                            &   & SRe$^2$L & Minimax & RDED & \textbf{Ours} & SRe$^2$L  & Minimax & RDED & \textbf{Ours} \\ \midrule
\multirow{3}{*}{ResNet-18}  & 1  & 13.3 $\pm$ 0.5 & 19.9 $\pm$ 0.2  & 20.8 $\pm$ 1.2 & \textbf{21.1 $\pm$ 0.6}   & 19.1 $\pm$ 1.1 & 31.8 $\pm$ 0.6   & 35.8 $\pm$ 1.0 & \textbf{40.6 $\pm$ 0.6}   \\
& 10 & 20.2 $\pm$ 0.2 & 40.1 $\pm$ 1.0  & 38.5 $\pm$ 2.1 & \textbf{45.6 $\pm$ 1.4} & 29.4 $\pm$ 3.0 & 61.4 $\pm$ 0.7 & 61.4 $\pm$ 0.4 & \textbf{65.0 $\pm$ 0.7} \\
& 50  & 23.3 $\pm$ 0.3 & 67.0 $\pm$ 1.8  & 68.5 $\pm$ 0.7 & \textbf{68.9 $\pm$ 1.1}   & 40.9 $\pm$ 0.3 & 84.1 $\pm$ 0.2   & 80.4 $\pm$ 0.4 & \textbf{84.5 $\pm$ 0.6} \\ \midrule
\multirow{3}{*}{ResNet-50}  & 1  & 14.9 $\pm$ 0.6   & 19.5 $\pm$ 0.6  & 14.5 $\pm$ 1.2   & \textbf{20.6 $\pm$ 1.6}   & 17.5 $\pm$ 3.4   & 27.9 $\pm$ 0.2   & 32.3 $\pm$ 0.8   & \textbf{33.5 $\pm$ 2.1} \\
& 10 & 17.3 $\pm$ 1.7 & 37.3 $\pm$ 1.1  & 29.9 $\pm$ 2.2 & \textbf{40.1 $\pm$ 0.1} & 49.8 $\pm$ 2.1 & 66.4 $\pm$ 0.4 & 63.9 $\pm$ 0.7   & \textbf{67.5 $\pm$ 0.8} \\
& 50 & 24.8 $\pm$ 0.7 & 64.3 $\pm$ 0.9  & 67.8 $\pm$ 0.3 & \textbf{68.2 $\pm$ 1.1} & 71.2 $\pm$ 0.3 & 77.1 $\pm$ 0.7 & 78.0 $\pm$ 0.4   & \textbf{82.7 $\pm$ 0.3} \\ \midrule
\multirow{3}{*}{ResNet-101} & 1  & 13.4 $\pm$ 0.1 & 17.7 $\pm$ 0.9  & 19.6 $\pm$ 1.8 & \textbf{21.2 $\pm$ 1.7} & 15.8 $\pm$ 0.6 & 24.5 $\pm$ 0.1 & 25.1 $\pm$ 2.7 & \textbf{32.7 $\pm$ 2.5} \\
& 10 & 17.7 $\pm$ 0.9 & 34.2 $\pm$  1.7 & 31.3 $\pm$ 1.3 & \textbf{36.5 $\pm$ 1.4} & 23.4 $\pm$ 0.8 & 55.4 $\pm$ 4.5 & 54.0 $\pm$ 0.4 & \textbf{66.3 $\pm$ 1.3} \\
& 50 & 21.2 $\pm$ 0.2 & 62.7 $\pm$ 1.6  & 59.1 $\pm$ 0.7 & \textbf{63.1 $\pm$ 1.3} & 36.5 $\pm$ 0.7 & 77.4 $\pm$ 0.8 & 75.0 $\pm$ 1.2 & \textbf{81.7 $\pm$ 1.0} \\ \bottomrule
\end{tabular}
}
\vspace{-8pt}
\caption{\textbf{Performance comparison (\%) with the SOTA methods over ImageNet subsets.} We adopt the evaluation paradigms from \cite{minimax} and \cite{rded}, reporting the higher accuracy achieved between the two. Best results are marked in \textbf{bold}.}
\label{tab:subet_acc}
\vspace{-12pt}
\end{table*}

\subsection{Experimental Settings}
\noindent\textbf{Dataset and Evaluation Settings.} For the goal of more practical and realistic application, we experiment on ImageNet \cite{imagenet} and its subsets, including ImageWoof \cite{imagewoof}, ImageNette \cite{imagenette} and ImageNet-100 \cite{imagenet100}. ImageWoof is a challenging subset containing 10 different dog breeds, while ImageNette is a simpler subset with 10 easily distinguishable categories. ImageNet-100 includes 100 randomly selected classes from the broader ImageNet. Aiming at higher compression ratios with generalizability, we adopt three images-per-class (IPC) settings of 1, 10, and 50 for each dataset. 

Currently, two evaluation paradigms exist for distilled dataset assessment: one based on hard labels \cite{minimax} and the other on soft labels via knowledge distillation \cite{rded}. While the latter often yields superior performance, it heavily depends on the expert model's accuracy and may occasionally fail (Appendix \ref{sec:eval_comp}). \textit{To ensure fairness and robustness, we evaluate \textbf{both} approaches and report the best result for each method.} The distilled datasets are tested across various model architectures, including ResNet-18 \cite{resnet}, ResNet-50, ResNet-101, EfficientNet-B0 \cite{efficientnet}, and MobileNet-V2 \cite{mobilenet}. To ensure stability, all experiments are repeated three times.

\noindent\textbf{Baselines.} We primarily compare our method with other diffusion-based dataset distillation methods and SOTA baselines, including SRe$^2$L \cite{sre2l}, Minimax Diffusion \cite{minimax}, D$^4$M \cite{d4m} and RDED \cite{rded}. We also include comparisons with methods such as G-VBSM \cite{gvbsm}, EDC \cite{edc}, IGD \cite{igd}, DATM \cite{datm}, and PAD \cite{pad} in Appendix \ref{sec:exp_supplement}. SRe$^2$L pioneered efficient scaling to ImageNet-1K and outperforms other matching-based methods like MTT \cite{mtt} and TESLA \cite{tesla}. Minimax Diffusion and D$^4$M utilized pre-trained diffusion models for efficient distillation, with \cite{minimax} applying a minimax criterion to enhance sample representativeness and diversity, while \cite{d4m} clusters image latents into category prototypes to incorporate label information. RDED, on the other hand, uses the original training set by selecting and grouping the most informative image crops.

\noindent\textbf{Implementation Details.} We adopt the pre-trained Diffusion Transformer (DiT) \cite{dit} with 256*256 resolution as the model backbone. Image sampling is performed with 50 denoising steps using a fixed random seed. For the image refinement stage, we adopt the Adam \cite{adam} optimizer with a learning rate of 0.0006, set $\lambda=10$ for the regularization hyperparameter, and train each image for 100 iterations. During training, we fix the input noise for each image and randomly sample a time step at each iteration. All experiments can be conducted on a single RTX A6000 GPU.

\subsection{Results and Discussions}
We present the comparison of distillation performance in \cref{tab:subet_acc} and \cref{tab:large_acc}. Our method is independent of the evaluation model, allowing us to use the same distilled dataset for assessments across ResNets of varying depths.

\noindent\textbf{ImageWoof and ImageNette}: As two smaller subsets of ImageNet, these tasks capture different aspects of the ImageNet distribution. ImageWoof includes various classes within a single species, testing the representativeness of the distilled dataset. In contrast, ImageNette, an easier dataset with distinct categories, assesses the diversity of the distilled dataset. \cref{tab:subet_acc} demonstrates that on these two datasets, SRe$^2$L performs sub-optimally, while Minimax Diffusion achieves results comparable to RDED. Our method surpasses the best baselines, achieving an average accuracy improvement of 1.6\% across all settings for ImageWoof and 4.3\% for ImageNette.


\noindent\textbf{ImageNet-100 and ImageNet-1K}: We further scale up to evaluations on 100 and 1,000 classes, with results shown in \cref{tab:large_acc}. For ImageNet-100, our method achieves an average accuracy improvement of 1.8\% across all settings. On ImageNet-1K, we include D$^4$M \cite{d4m} for comparison, using the reported accuracies and assuming zero deviation, as repeated results were not provided in the original paper. Our method consistently outperforms other methods in all settings, yielding an average accuracy improvement of 1.5\%.

\begin{table*}[t]
\centering
\small
\adjustbox{width=1.0\textwidth}{
\begin{tabular}{cc|cccc|ccccc}
\toprule
 &  & \multicolumn{4}{c|}{ImageNet-100}  & \multicolumn{5}{c}{ImageNet-1K}  \\
\multirow{-2}{*}{Settings}   & \multirow{-2}{*}{IPC} & SRe$^2$L   & Minimax  & RDED  & \textbf{Ours}   & SRe$^2$L & Minimax & D$^4$M  & RDED  & \textbf{Ours}   \\ \midrule
 & 1  & 3.0$\pm$0.3  & 7.3$\pm$0.1  & 8.1$\pm$0.3  & \textbf{8.8$\pm$0.4} & 0.1$\pm$0.1  & 5.9$\pm$0.2 & -  & 6.6$\pm$0.2& \textbf{7.1$\pm$0.1}  \\
 & 10 & 9.5$\pm$0.4  & 32.0$\pm$1.0 & 36.0$\pm$0.3 & \textbf{36.6$\pm$0.2} & 21.3$\pm$0.6 & 44.3$\pm$0.5  & 27.9$\pm$0.0 & 42.0$\pm$0.1 & \textbf{46.1$\pm$0.2} \\
\multirow{-3}{*}{ResNet-18}  & 50 & 27.0$\pm$0.4 & 63.9$\pm$0.1 & 61.6$\pm$0.1 & \textbf{68.0$\pm$0.5} & 46.8$\pm$0.2 & 58.6$\pm$0.3  & 55.2$\pm$0.0 & 56.5$\pm$0.1 & \textbf{60.0$\pm$0.0}  \\ \midrule
 & 1 & 1.5$\pm$0.0  & 6.8$\pm$0.5  & 6.5$\pm$0.3    & \textbf{7.3$\pm$0.4} & 1.0$\pm$0.0 & 5.1$\pm$0.5   & -  & 5.4$\pm$0.2    & \textbf{7.0$\pm$0.4} \\
 & 10 & 3.4$\pm$0.1  & 30.8$\pm$0.4 & 29.5$\pm$0.7   & \textbf{35.0$\pm$0.6} & 28.4$\pm$0.1   & 49.7$\pm$0.8  & 33.5$\pm$0.0 & 43.6$\pm$0.5   & \textbf{53.0$\pm$0.2}  \\
\multirow{-3}{*}{ResNet-50}  & 50  & 10.8$\pm$0.3   & 67.4$\pm$0.3 & 68.8$\pm$0.2   & \textbf{70.1$\pm$0.1}  & 55.6$\pm$0.3   & 64.8$\pm$0.1  & 62.4$\pm$0.0 & 64.6$\pm$0.1   & \textbf{65.5$\pm$0.1}  \\ \midrule
 & 1  & 2.1$\pm$0.1  & 5.4$\pm$0.6  & 6.1$\pm$0.8  & \textbf{6.6$\pm$0.4}  & 0.6$\pm$0.1  & 4.0$\pm$0.5   & -   & 5.9$\pm$0.4  & \textbf{6.0$\pm$0.4}   \\
  & 10 & 6.4$\pm$0.1  & 29.2$\pm$1.0 & 33.9$\pm$0.1 & \textbf{34.5$\pm$0.4}& 30.9$\pm$0.1 & 46.9$\pm$1.3  & 34.2$\pm$0.0 & 48.3$\pm$1.0 & \textbf{52.2$\pm$1.1}  \\
\multirow{-3}{*}{ResNet-101} & 50   & 25.7$\pm$0.3 & 67.4$\pm$0.6 & 66.0$\pm$0.6 & \textbf{70.8$\pm$0.2} & 60.8$\pm$0.5 & 65.6$\pm$0.1  & 63.4$\pm$0.0 & 61.2$\pm$0.4 & \textbf{66.2$\pm$0.1}  \\ \hline
\end{tabular}
}
\vspace{-8pt}
\caption{\textbf{Performance comparison (\%) over ImageNet-100 and ImageNet-1K.} We adopt the evaluation paradigms from \cite{minimax} and \cite{rded}, reporting the higher accuracy achieved. '-' indicates missing results from the original paper. Best results are marked in \textbf{bold}.}
\label{tab:large_acc}
\vspace{-16pt}
\end{table*}

\subsection{Ablation Studies}
We conduct extensive ablation studies to evaluate the effectiveness of the different components in our method, and explore the impact of different hyperparameter choices.

\noindent\textbf{Component Analysis.} We analyze the effectiveness of each component in our method using ResNet-18 as the model for evaluation. \cref{tab:component} demonstrates that both components individually improve upon the baseline, and their combination yields even larger performance gains. Furthermore, the comparable accuracy improvements provided by each component indicate their equal importance.

\begin{table}[h]
\centering
\small
\adjustbox{width=1.0\linewidth}{
\begin{tabular}{cc|cc|cc}
\toprule
OSS & CLO & \begin{tabular}[c]{@{}c@{}}Woof\\ IPC=10\end{tabular} & \begin{tabular}[c]{@{}c@{}}Woof\\ IPC=50\end{tabular} & \begin{tabular}[c]{@{}c@{}}Nette\\ IPC=10\end{tabular} & \begin{tabular}[c]{@{}c@{}}Nette\\ IPC=50\end{tabular} \\ \midrule
- & - & 38.7$\pm$1.1 & 66.1$\pm$0.8 & 61.7$\pm$1.7 & 82.0$\pm$1.1 \\
\CheckmarkBold & - & 42.6$\pm$1.1   & 68.5$\pm$0.9  & 63.7$\pm$0.6  & 83.5$\pm$1.1   \\
-  & \CheckmarkBold  & 42.1$\pm$1.2   & 67.9$\pm$0.8   & 64.2$\pm$2.1   & 83.9$\pm$0.7  \\
\CheckmarkBold & \CheckmarkBold  & 45.6$\pm$1.4  & 68.7$\pm$0.5  & 65.0$\pm$0.7 & 84.5$\pm$0.6 \\ \bottomrule
\end{tabular}
}
\vspace{-8pt}
\caption{Accuracy (\%) comparison of different components.}
\label{tab:component}
\vspace{-16pt}
\end{table}

\begin{figure*}[t]
\begin{center}
\centerline{\includegraphics[trim=10 55 5 40,clip, width=1.0\textwidth]{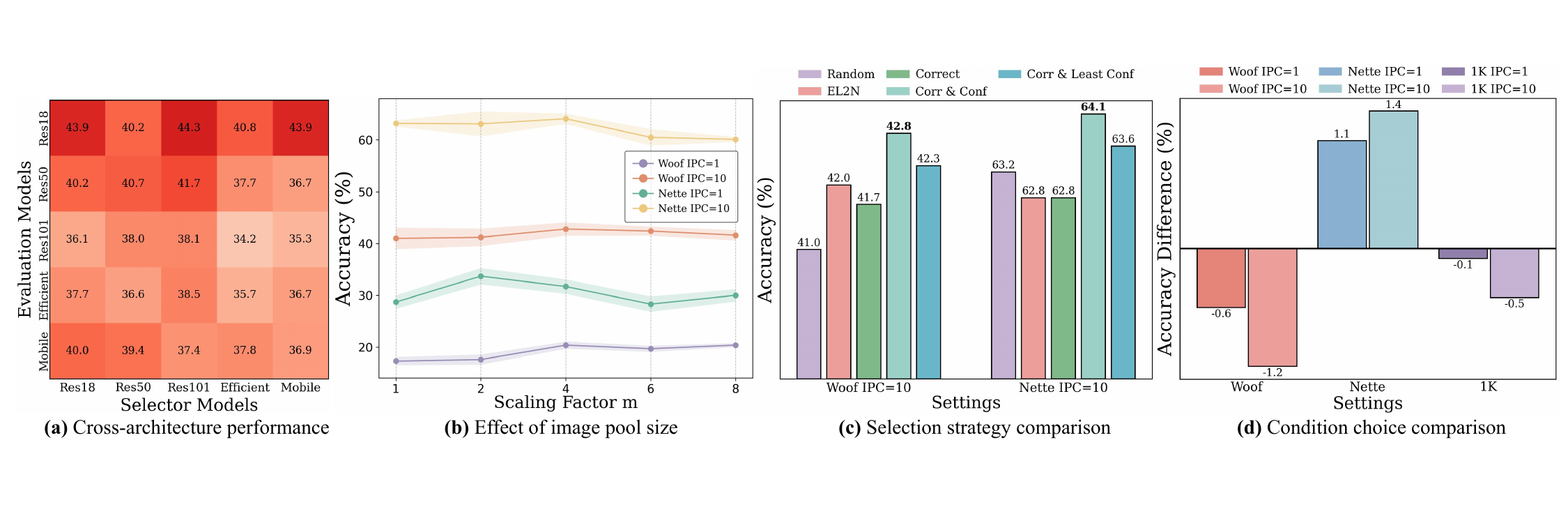}}
\vspace{-10pt}
\caption{\textbf{Ablation studies on different components.} \textbf{(a)} illustrates the cross-architecture performance for selector and evaluation models, showing that a lightweight ResNet-18 generally leads to better performance in most cases. \textbf{(b)} examines the effect of the scaling factor, showing that a larger pool size is not always optimal. As a result, we adopt a pool size of 2$\times$IPC or 4$\times$IPC. \textbf{(c)} compares different selection strategies, concluding that prioritizing correct and more confident samples generally leads to better performance. \textbf{(d)} calculates the accuracy difference when applying various conditions for latent optimization, suggesting that true labels should be used for easier class samples, while unconditional labels are more suitable for harder class samples.}
\label{fig:ablation}
\end{center}
\vspace{-30pt}
\end{figure*}

\noindent\textbf{Selection Protocols.} We first examine the effects of selection pool size and selection strategy. As shown in \cref{fig:ablation}(b), an optimal pool size for sample selection typically ranges around 2$\times$IPC or 4$\times$IPC. Interestingly, selecting from a larger set of generated samples does not necessarily improve performance. We suspect this is due to two factors: first, while a larger pool provides more diversity, selecting from it may reduce representativeness \cite{minimax}; second, although our empirical results (as will be discussed later) support the effectiveness of the current selection criterion, it is not guaranteed to consistently identify the optimal samples. Developing more advanced and complex selection protocols may address these issues, but is beyond the scope of this study.

For the selection criterion, we evaluate several strategies: random sampling, EL2N score \cite{el2n} based selection, selecting only correct samples, selecting correct and more confident samples, and selecting correct but least confident samples. The EL2N score measures sample difficulty via probability output, with lower scores indicating samples that are easier to discriminate. As shown in \cref{fig:ablation}(c), selecting correct and more confident samples generally yields the best performance across most cases. However, random sampling performs best on ImageNette with 50 IPC, likely because the diffusion model is well-trained on ImageNette classes, preserving both representativeness and diversity. Random selection effectively covers the modeled distribution, whereas probability-based selection protocols may introduce unintended bias into the distilled dataset.

\noindent\textbf{Input condition comparison.} We further analyze the choice of $\mathbf{\hat{c}}$ in \cref{eq:condition}. In \cref{fig:ablation}(d), we present the performance difference, calculated as
$Acc(\text{true}) - Acc(\text{NULL})$,
when varying only the conditions used for latent optimization. The results suggest that for more challenging tasks such as ImageWoof and ImageNet-1K, unconditional labels yield better performance, while for easier tasks, such as ImageNette, using true class labels is preferable.

\noindent\textbf{Effect of regularization.}
Lastly, we examine the effect of different regularization forms in \cref{eq:objective}. We compare no regularization, $L_1$, $L_2$, and $L_\infty$ norms, tuning the hyperparameter $\lambda$ from $(10, 1, 0.1, 0.01)$ to achieve the best performance. \cref{tab:norm} presents the results, where using regularization most of the time improves performance and the $L_\infty$ norm performs the best. We attribute this to the $L_\infty$ norm’s ability to enhance the robustness of generated latent representations, aiding to prevent the learning process from producing outliers or extreme values.

\begin{table}[t]
\centering
\small
\adjustbox{width=1.0\linewidth}{
\begin{tabular}{c|cccc}
\toprule
$L_p$ norm & None &  $L_1$    & $L_2$    & $L_\infty$ \\ \midrule
ImageWoof (IPC=10) & 42.7$\pm$0.7 & 43.4$\pm$2.2 & 44.1$\pm$0.7 & \textbf{44.4$\pm$0.2}   \\
ImageNette (IPC=10) & 62.2$\pm$1.4 & 61.3$\pm$0.3 & 62.5$\pm$1.0 & \textbf{63.5$\pm$0.8}   \\ \bottomrule
\end{tabular}
}
\vspace{-8pt}
\caption{Accuracy (\%) comparison of different regularization.}
\label{tab:norm}
\vspace{-16pt}
\end{table}

\subsection{Cross-Architecture Performance}
We evaluate and analyze two types of cross-architecture performances. The first is the conventional comparison between different evaluation models when trained on the same synthetic dataset. Since diffusion-based methods do not rely on the evaluation model during distillation, \cref{tab:subet_acc} and \cref{tab:large_acc} themselves already show these comparison results, demonstrating the strong cross-architecture ability.

The second type of evaluation involves comparing performance across different combinations of selectors and evaluation models. As shown in \cref{fig:ablation}(a), the models included from left to right, top to bottom, are ResNet-18, ResNet-50, ResNet-101, EfficientNet-B0, MobileNet-V2, respectively. We observe that using stronger selectors does not consistently improve performance, and the lightweight ResNet-18 is sufficient for achieving good results efficiently. Therefore, we use the pre-trained ResNet-18 as the sample selector in all our experiments.

\begin{figure}
\begin{center}
\centerline{\includegraphics[trim=155 20 135 20,clip, width=1.0\linewidth]{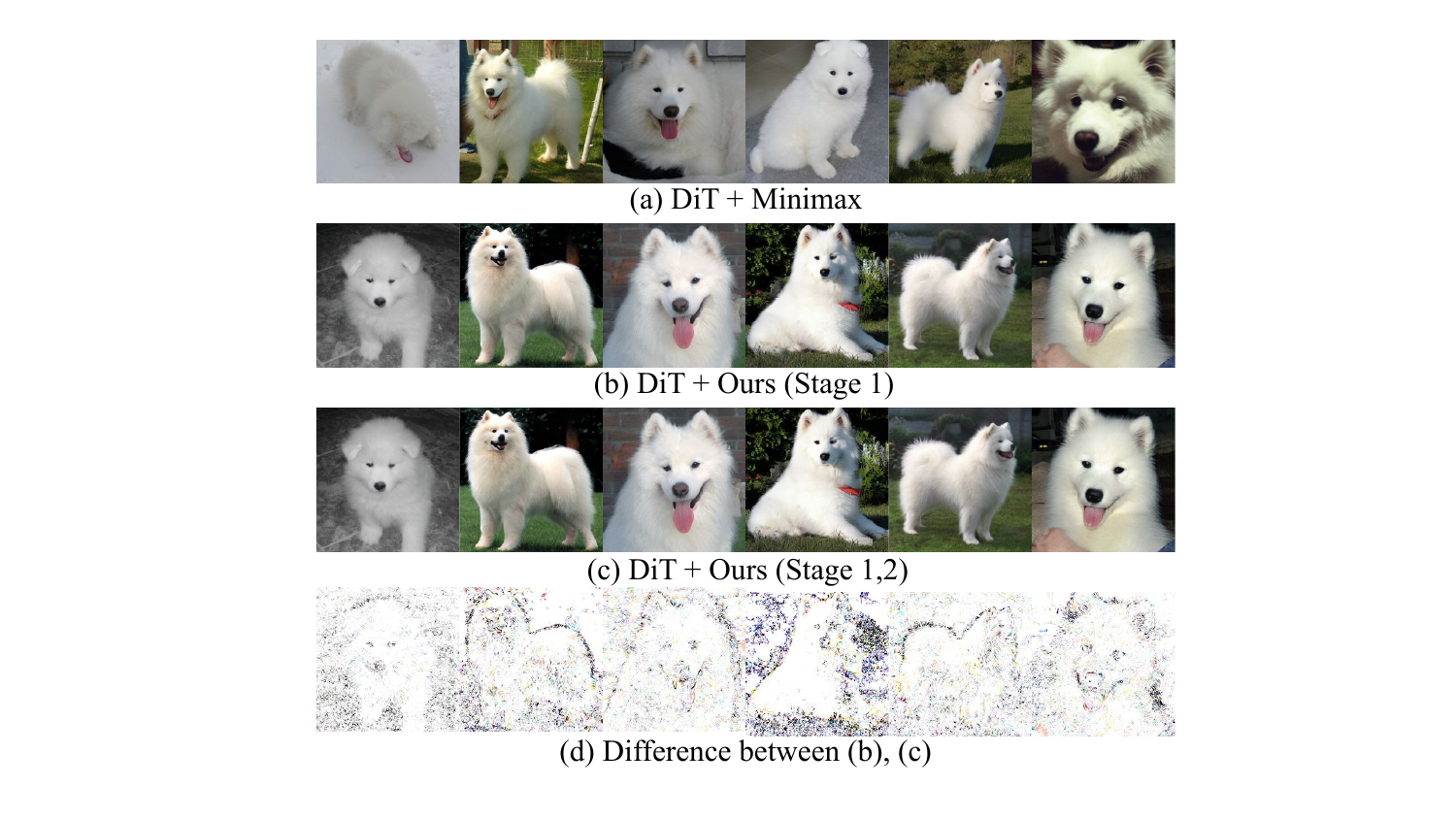}}
\vspace{-10pt}
\caption{\textbf{Visualization of distilled images.} We compare the distilled images from Minimax Diffusion with the two stages of our method, also highlighting the differences introduced by latent optimization, where white color denotes unchanged regions.}
\label{fig:examples}
\end{center}
\vspace{-30pt}
\end{figure}

\subsection{Visualization Comparison}
We visualize obtained samples using the same random seeds in \cref{fig:examples}, qualitatively revealing that our method produces more discriminative and visually refined images. Interestingly, latent optimization introduces imperceptible changes to the human eye. By examining the absolute differences between images before and after latent optimization, we find that the optimization primarily refines object boundaries and background details, suggesting that images might be enhanced by emphasized semantic features and improved robustness against adversarial pixels \cite{robust_classification}. Appendix \ref{sec:visual} provides more visualizations and analysis.

\subsection{Extending to Different Backbones}
\noindent \textbf{Integration with MAR.} As discussed in \cref{sec:mar}, we extend our method to masked autoregressive models \cite{mar}, using MAR-Base with 32 autoregressive steps and 100 denoising steps per autoregressive step. From a 2$\times$IPC image pool, we select the most confident correct samples. For latent optimization, we apply a 0.25 max masking ratio, train for 100 steps with a 0.0001 learning rate, and use $L_\infty$ regularization.
\cref{tab:backbone} shows that MAR alone is a strong baseline on ImageNet subsets, and combining it with our method boosts accuracy by 3.3\% on ImageWoof and 2.0\% on ImageNette. Visualizations are in Appendix \ref{sec:visual}. 

\begin{table}[t]
\centering
\small
\adjustbox{width=1.0\linewidth}{
\begin{tabular}{c|cc|cc}
\toprule
Method & \begin{tabular}[c]{@{}c@{}}Woof\\ IPC=10\end{tabular} & \begin{tabular}[c]{@{}c@{}}Woof\\ IPC=50\end{tabular} & \begin{tabular}[c]{@{}c@{}}Nette\\ IPC=10\end{tabular} & \begin{tabular}[c]{@{}c@{}}Nette\\ IPC=50\end{tabular} \\ \midrule
MAR \cite{mar} & 38.6$\pm$0.4  & 69.1$\pm$0.2 & 68.8$\pm$1.8  & 86.3$\pm$0.4 \\
+ CaO$_2$ & \textbf{43.9$\pm$0.8} & \textbf{70.3$\pm$0.1} & \textbf{71.8$\pm$1.6} & \textbf{87.3$\pm$0.1} \\ \midrule
Minimax \cite{minimax} & 40.1$\pm$1.0 & 67.0$\pm$1.8 & 61.4$\pm$0.7 & 83.9$\pm$0.2 \\
+ CaO$_2$ & \textbf{45.7$\pm$1.5} & \textbf{70.0$\pm$1.4} & \textbf{64.1$\pm$1.8} & \textbf{85.1$\pm$1.0} \\ \bottomrule
\end{tabular}
}
\vspace{-8pt}
\caption{Performance (\%) of integration with different backbones.}
\label{tab:backbone}
\vspace{-12pt}
\end{table}

\noindent\textbf{Integration with Minimax Diffusion.} We further integrate our method with Minimax Diffusion \cite{minimax}, a finetuned diffusion transformer that improves sample quality without altering the model architecture. Using a 2$\times$IPC image pool, $L_\infty$ regularization, and true label guidance (except zero embeddings for ImageWoof at IPC=10), \cref{tab:backbone} shows that we further improve the performance with an average improvement of 4.3\% on ImageWoof and 2.0\% on ImageNette. Visualizations are available in Appendix \ref{sec:visual}.
In summary, CaO$_2$ serves as an effective plug-and-play module for both diffusion models with diverse training objectives and masked autoregressive models with distinct generation schemes.

\section{Conclusion}
We propose CaO$_2$, an efficient two-stage framework for large-scale dataset distillation, addressing two key inconsistencies in diffusion-based dataset distillation: Objective Inconsistency and Condition Inconsistency. Our method achieves state-of-the-art performance on ImageNet and its subsets across various evaluation settings. Additionally, CaO$_2$ can be seamlessly integrated with different model backbones to enhance distillation performance, showcasing its versatility as a highly effective plug-and-play approach.

\noindent\textbf{Acknowledgments:} This research is supported by NSF IIS-2525840, CNS-2432534, ECCS-2514574, NIH 1RF1MH133764-01 and Cisco Research unrestricted gift. This article solely reflects opinions and conclusions of authors and not funding agencies.
{
    \small
    \bibliographystyle{ieeenat_fullname}
    \bibliography{main}

\begin{thebibliography}{48}
\providecommand{\natexlab}[1]{#1}
\providecommand{\url}[1]{\texttt{#1}}
\expandafter\ifx\csname urlstyle\endcsname\relax
  \providecommand{\doi}[1]{doi: #1}\else
  \providecommand{\doi}{doi: \begingroup \urlstyle{rm}\Url}\fi

\bibitem[Bengio et~al.(2009)Bengio, Louradour, Collobert, and Weston]{curriculum}
Yoshua Bengio, J{\'e}r{\^o}me Louradour, Ronan Collobert, and Jason Weston.
\newblock Curriculum learning.
\newblock In \emph{International Conference on Machine Learning}, 2009.

\bibitem[Cazenavette et~al.(2022)Cazenavette, Wang, Torralba, Efros, and Zhu]{mtt}
George Cazenavette, Tongzhou Wang, Antonio Torralba, Alexei~A. Efros, and Jun-Yan Zhu.
\newblock Dataset distillation by matching training trajectories.
\newblock In \emph{Proceedings of the IEEE/CVF Conference on Computer Vision and Pattern Recognition}, 2022.

\bibitem[Cazenavette et~al.(2023)Cazenavette, Wang, Torralba, Efros, and Zhu]{glad}
George Cazenavette, Tongzhou Wang, Antonio Torralba, Alexei~A. Efros, and Jun-Yan Zhu.
\newblock Generalizing dataset distillation via deep generative prior.
\newblock In \emph{Proceedings of the IEEE/CVF Conference on Computer Vision and Pattern Recognition}, 2023.

\bibitem[Chen et~al.(2023)Chen, Dong, Wang, Yang, Duan, Su, and Zhu]{robust_classification}
Huanran Chen, Yinpeng Dong, Zhengyi Wang, Xiao Yang, Chengqi Duan, Hang Su, and Jun Zhu.
\newblock Robust classification via a single diffusion model.
\newblock \emph{arXiv preprint arXiv:2305.15241}, 2023.

\bibitem[Chen et~al.(2025)Chen, Du, Huang, Wang, Zhang, and Wang]{igd}
Mingyang Chen, Jiawei Du, Bo Huang, Yi Wang, Xiaobo Zhang, and Wei Wang.
\newblock Influence-guided diffusion for dataset distillation.
\newblock In \emph{The Thirteenth International Conference on Learning Representations}, 2025.

\bibitem[Cui et~al.(2023)Cui, Wang, Si, and Hsieh]{tesla}
Justin Cui, Ruochen Wang, Si Si, and Cho-Jui Hsieh.
\newblock Scaling up dataset distillation to imagenet-1k with constant memory.
\newblock In \emph{International Conference on Machine Learning}, pages 6565--6590. PMLR, 2023.

\bibitem[Deng et~al.(2009)Deng, Dong, Socher, Li, Li, and Fei-Fei]{imagenet}
Jia Deng, Wei Dong, Richard Socher, Li-Jia Li, Kai Li, and Li Fei-Fei.
\newblock Imagenet: A large-scale hierarchical image database.
\newblock In \emph{2009 IEEE conference on computer vision and pattern recognition}, pages 248--255. Ieee, 2009.

\bibitem[Du et~al.(2023)Du, Jiang, Tan, Zhou, and Li]{ftd}
Jiawei Du, Yidi Jiang, Vincent~YF Tan, Joey~Tianyi Zhou, and Haizhou Li.
\newblock Minimizing the accumulated trajectory error to improve dataset distillation.
\newblock In \emph{Proceedings of the IEEE/CVF conference on computer vision and pattern recognition}, pages 3749--3758, 2023.

\bibitem[Gu et~al.(2024)Gu, Vahidian, Kungurtsev, Wang, Jiang, You, and Chen]{minimax}
Jianyang Gu, Saeed Vahidian, Vyacheslav Kungurtsev, Haonan Wang, Wei Jiang, Yang You, and Yiran Chen.
\newblock Efficient dataset distillation via minimax diffusion.
\newblock In \emph{Proceedings of the IEEE/CVF Conference on Computer Vision and Pattern Recognition (CVPR)}, 2024.

\bibitem[Guo et~al.(2024)Guo, Wang, Cazenavette, Li, Zhang, and You]{datm}
Ziyao Guo, Kai Wang, George Cazenavette, Hui Li, Kaipeng Zhang, and Yang You.
\newblock Towards lossless dataset distillation via difficulty-aligned trajectory matching.
\newblock In \emph{The Twelfth International Conference on Learning Representations}, 2024.

\bibitem[He et~al.(2015)He, Zhang, Ren, and Sun]{resnet}
Kaiming He, Xiangyu Zhang, Shaoqing Ren, and Jian Sun.
\newblock Deep residual learning for image recognition, 2015.

\bibitem[He et~al.(2023)He, Xiao, and Zhou]{yoco}
Yang He, Lingao Xiao, and Joey~Tianyi Zhou.
\newblock You only condense once: Two rules for pruning condensed datasets, 2023.

\bibitem[Howard et~al.(2017)Howard, Zhu, Chen, Kalenichenko, Wang, Weyand, Andreetto, and Adam]{mobilenet}
Andrew~G. Howard, Menglong Zhu, Bo Chen, Dmitry Kalenichenko, Weijun Wang, Tobias Weyand, Marco Andreetto, and Hartwig Adam.
\newblock Mobilenets: Efficient convolutional neural networks for mobile vision applications, 2017.

\bibitem[Howard(2019{\natexlab{a}})]{imagenette}
Jeremy Howard.
\newblock Imagenette: A smaller subset of 10 easily classified classes from imagenet, 2019{\natexlab{a}}.

\bibitem[Howard(2019{\natexlab{b}})]{imagewoof}
Jeremy Howard.
\newblock Imagewoof: a subset of 10 classes from imagenet that aren't so easy to classify, 2019{\natexlab{b}}.

\bibitem[Kim et~al.(2022)Kim, Kim, Oh, Yun, Song, Jeong, Ha, and Song]{idc}
Jang-Hyun Kim, Jinuk Kim, Seong~Joon Oh, Sangdoo Yun, Hwanjun Song, Joonhyun Jeong, Jung-Woo Ha, and Hyun~Oh Song.
\newblock Dataset condensation via efficient synthetic-data parameterization.
\newblock In \emph{International Conference on Machine Learning (ICML)}, 2022.

\bibitem[Kingma and Ba(2017)]{adam}
Diederik~P. Kingma and Jimmy Ba.
\newblock Adam: A method for stochastic optimization, 2017.

\bibitem[Li et~al.(2023)Li, Prabhudesai, Duggal, Brown, and Pathak]{diffusion_classifier}
Alexander~C. Li, Mihir Prabhudesai, Shivam Duggal, Ellis Brown, and Deepak Pathak.
\newblock Your diffusion model is secretly a zero-shot classifier.
\newblock In \emph{Proceedings of the IEEE/CVF International Conference on Computer Vision (ICCV)}, pages 2206--2217, 2023.

\bibitem[Li et~al.(2024{\natexlab{a}})Li, Tian, Li, Deng, and He]{mar}
Tianhong Li, Yonglong Tian, He Li, Mingyang Deng, and Kaiming He.
\newblock Autoregressive image generation without vector quantization.
\newblock \emph{arXiv preprint arXiv:2406.11838}, 2024{\natexlab{a}}.

\bibitem[Li et~al.(2024{\natexlab{b}})Li, Guo, Zhao, Zhang, Cheng, Khaki, Zhang, Sajedi, Plataniotis, Wang, and You]{dd_sample_diff}
Zekai Li, Ziyao Guo, Wangbo Zhao, Tianle Zhang, Zhi-Qi Cheng, Samir Khaki, Kaipeng Zhang, Ahmad Sajedi, Konstantinos~N Plataniotis, Kai Wang, and Yang You.
\newblock Prioritize alignment in dataset distillation, 2024{\natexlab{b}}.

\bibitem[Li et~al.(2024{\natexlab{c}})Li, Guo, Zhao, Zhang, Cheng, Khaki, Zhang, Sajedi, Plataniotis, Wang, and You]{pad}
Zekai Li, Ziyao Guo, Wangbo Zhao, Tianle Zhang, Zhi-Qi Cheng, Samir Khaki, Kaipeng Zhang, Ahmad Sajedi, Konstantinos~N Plataniotis, Kai Wang, and Yang You.
\newblock Prioritize alignment in dataset distillation, 2024{\natexlab{c}}.

\bibitem[Li et~al.(2024{\natexlab{d}})Li, Zhong, Liang, Zhou, Shi, Wang, Zhao, Zhao, Wang, Qin, Liu, Zhang, Zhou, Zhu, Wang, Li, Zhang, Liu, Huang, Lyu, Lv, Jin, Akata, Gu, Vedantam, Shou, Deng, Yan, Shang, Cazenavette, Wu, Cui, Chen, Yao, Kellis, Plataniotis, Zhao, Wang, You, and Wang]{ddranking}
Zekai Li, Xinhao Zhong, Zhiyuan Liang, Yuhao Zhou, Mingjia Shi, Ziqiao Wang, Wangbo Zhao, Xuanlei Zhao, Haonan Wang, Ziheng Qin, Dai Liu, Kaipeng Zhang, Tianyi Zhou, Zheng Zhu, Kun Wang, Guang Li, Junhao Zhang, Jiawei Liu, Yiran Huang, Lingjuan Lyu, Jiancheng Lv, Yaochu Jin, Zeynep Akata, Jindong Gu, Rama Vedantam, Mike Shou, Zhiwei Deng, Yan Yan, Yuzhang Shang, George Cazenavette, Xindi Wu, Justin Cui, Tianlong Chen, Angela Yao, Manolis Kellis, Konstantinos~N. Plataniotis, Bo Zhao, Zhangyang Wang, Yang You, and Kai Wang.
\newblock Dd-ranking: Rethinking the evaluation of dataset distillation.
\newblock GitHub repository, 2024{\natexlab{d}}.

\bibitem[Liu et~al.(2019)Liu, Simonyan, and Yang]{darts}
Hanxiao Liu, Karen Simonyan, and Yiming Yang.
\newblock Darts: Differentiable architecture search, 2019.

\bibitem[Liu et~al.(2023)Liu, Gu, Wang, Zhu, Jiang, and You]{dream}
Yanqing Liu, Jianyang Gu, Kai Wang, Zheng Zhu, Wei Jiang, and Yang You.
\newblock Dream: Efficient dataset distillation by representative matching, 2023.

\bibitem[Masarczyk and Tautkute(2020)]{continual}
Wojciech Masarczyk and Ivona Tautkute.
\newblock Reducing catastrophic forgetting with learning on synthetic data, 2020.

\bibitem[Moser et~al.(2024)Moser, Raue, Palacio, Frolov, and Dengel]{ld3m}
Brian~B. Moser, Federico Raue, Sebastian Palacio, Stanislav Frolov, and Andreas Dengel.
\newblock Latent dataset distillation with diffusion models, 2024.

\bibitem[Paul et~al.(2023)Paul, Ganguli, and Dziugaite]{el2n}
Mansheej Paul, Surya Ganguli, and Gintare~Karolina Dziugaite.
\newblock Deep learning on a data diet: Finding important examples early in training, 2023.

\bibitem[Peebles and Xie(2022)]{dit}
William Peebles and Saining Xie.
\newblock Scalable diffusion models with transformers.
\newblock \emph{arXiv preprint arXiv:2212.09748}, 2022.

\bibitem[Rombach et~al.(2021)Rombach, Blattmann, Lorenz, Esser, and Ommer]{sd}
Robin Rombach, Andreas Blattmann, Dominik Lorenz, Patrick Esser, and Björn Ommer.
\newblock High-resolution image synthesis with latent diffusion models, 2021.

\bibitem[Russakovsky et~al.(2015)Russakovsky, Deng, Su, Krause, Satheesh, Ma, Huang, Karpathy, Khosla, Bernstein, Berg, and Fei-Fei]{imagenet100}
Olga Russakovsky, Jia Deng, Hao Su, Jonathan Krause, Sanjeev Satheesh, Sean Ma, Zhiheng Huang, Andrej Karpathy, Aditya Khosla, Michael Bernstein, Alexander~C. Berg, and Li Fei-Fei.
\newblock Imagenet large scale visual recognition challenge, 2015.

\bibitem[Shao et~al.(2024)Shao, Yin, Zhou, Zhang, and Shen]{gvbsm}
Shitong Shao, Zeyuan Yin, Muxin Zhou, Xindong Zhang, and Zhiqiang Shen.
\newblock Generalized large-scale data condensation via various backbone and statistical matching, 2024.

\bibitem[Shao et~al.(2025)Shao, Zhou, Chen, and Shen]{edc}
Shitong Shao, Zikai Zhou, Huanran Chen, and Zhiqiang Shen.
\newblock Elucidating the design space of dataset condensation, 2025.

\bibitem[Su et~al.(2024)Su, Hou, Gao, Tian, and Tang]{d4m}
Duo Su, Junjie Hou, Weizhi Gao, Yingjie Tian, and Bowen Tang.
\newblock D{\textasciicircum}4m: Dataset distillation via disentangled diffusion model.
\newblock In \emph{Proceedings of the IEEE/CVF Conference on Computer Vision and Pattern Recognition (CVPR)}, pages 5809--5818, 2024.

\bibitem[Sun et~al.(2024)Sun, Shi, Yu, and Lin]{rded}
Peng Sun, Bei Shi, Daiwei Yu, and Tao Lin.
\newblock On the diversity and realism of distilled dataset: An efficient dataset distillation paradigm.
\newblock In \emph{Proceedings of the IEEE/CVF Conference on Computer Vision and Pattern Recognition (CVPR)}, 2024.

\bibitem[Tan and Le(2020)]{efficientnet}
Mingxing Tan and Quoc~V. Le.
\newblock Efficientnet: Rethinking model scaling for convolutional neural networks, 2020.

\bibitem[Wang et~al.(2022)Wang, Zhao, Peng, Zhu, Yang, Wang, Huang, Bilen, Wang, and You]{cafe}
Kai Wang, Bo Zhao, Xiangyu Peng, Zheng Zhu, Shuo Yang, Shuo Wang, Guan Huang, Hakan Bilen, Xinchao Wang, and Yang You.
\newblock Cafe: Learning to condense dataset by aligning features.
\newblock In \emph{Proceedings of the IEEE/CVF Conference on Computer Vision and Pattern Recognition}, pages 12196--12205, 2022.

\bibitem[Wang et~al.(2025)Wang, Yang, Liu, Sun, Hu, He, and Zhang]{ncfm}
Shaobo Wang, Yicun Yang, Zhiyuan Liu, Chenghao Sun, Xuming Hu, Conghui He, and Linfeng Zhang.
\newblock Dataset distillation with neural characteristic function: A minmax perspective, 2025.

\bibitem[Wang et~al.(2020)Wang, Zhu, Torralba, and Efros]{dd}
Tongzhou Wang, Jun-Yan Zhu, Antonio Torralba, and Alexei~A. Efros.
\newblock Dataset distillation, 2020.

\bibitem[Xu et~al.(2024)Xu, Li, Cui, Wang, Lu, Tai, and Tang]{bilp}
Yue Xu, Yong-Lu Li, Kaitong Cui, Ziyu Wang, Cewu Lu, Yu-Wing Tai, and Chi-Keung Tang.
\newblock Distill gold from massive ores: Bi-level data pruning towards efficient dataset distillation, 2024.

\bibitem[Yin and Shen(2023)]{cda}
Zeyuan Yin and Zhiqiang Shen.
\newblock Dataset distillation in large data era.
\newblock \emph{arXiv preprint arXiv:2311.18838}, 2023.

\bibitem[Yin et~al.(2023)Yin, Xing, and Shen]{sre2l}
Zeyuan Yin, Eric Xing, and Zhiqiang Shen.
\newblock Squeeze, recover and relabel: Dataset condensation at imagenet scale from a new perspective.
\newblock In \emph{Thirty-seventh Conference on Neural Information Processing Systems}, 2023.

\bibitem[Yu et~al.(2023)Yu, Liu, and Wang]{dd_survey}
Ruonan Yu, Songhua Liu, and Xinchao Wang.
\newblock Dataset distillation: A comprehensive review.
\newblock \emph{IEEE Transactions on Pattern Analysis and Machine Intelligence}, 2023.

\bibitem[Zhao and Bilen(2023)]{dm}
Bo Zhao and Hakan Bilen.
\newblock Dataset condensation with distribution matching.
\newblock In \emph{Proceedings of the IEEE/CVF Winter Conference on Applications of Computer Vision}, pages 6514--6523, 2023.

\bibitem[Zhao et~al.(2021)Zhao, Mopuri, and Bilen]{dc}
Bo Zhao, Konda~Reddy Mopuri, and Hakan Bilen.
\newblock Dataset condensation with gradient matching.
\newblock In \emph{International Conference on Learning Representations}, 2021.

\bibitem[Zhao et~al.(2023)Zhao, Li, Qin, and Yu]{idm}
Ganlong Zhao, Guanbin Li, Yipeng Qin, and Yizhou Yu.
\newblock Improved distribution matching for dataset condensation.
\newblock In \emph{Proceedings of the IEEE/CVF Conference on Computer Vision and Pattern Recognition}, pages 7856--7865, 2023.

\bibitem[Zhao et~al.(2024{\natexlab{a}})Zhao, Shang, Wu, and Yan]{dq}
Zhenghao Zhao, Yuzhang Shang, Junyi Wu, and Yan Yan.
\newblock Dataset quantization with active learning based adaptive sampling.
\newblock \emph{arXiv preprint arXiv:2407.07268}, 2024{\natexlab{a}}.

\bibitem[Zhao et~al.(2024{\natexlab{b}})Zhao, Wang, Shang, Wang, and Yan]{ltdd}
Zhenghao Zhao, Haoxuan Wang, Yuzhang Shang, Kai Wang, and Yan Yan.
\newblock Distilling long-tailed datasets, 2024{\natexlab{b}}.

\bibitem[Zhou et~al.(2024)Zhou, Yin, Shao, and Shen]{scdd}
Muxin Zhou, Zeyuan Yin, Shitong Shao, and Zhiqiang Shen.
\newblock Self-supervised dataset distillation: A good compression is all you need, 2024.

\end{thebibliography}
}

\clearpage
\setcounter{page}{1}
\maketitlesupplementary

\noindent The supplementary material is organized as follows: \cref{sec:mar_supplement} presents the process of integrating our method with MAR; \cref{sec:exp_supplement} includes more baseline comparisons and discussions; \cref{sec:ab_image} provides more ablation studies; \cref{sec:eval_comp} provide a more in-depth analysis of different evaluation paradigms; \cref{sec:visual} shows more examples of distilled images across different datasets; and \cref{sec:impact} discusses the limitations and broader impacts.

\section{Generalizing to MAR}
\label{sec:mar_supplement}
\begin{figure}[h]
\vspace{-15pt}
\begin{center}
\centerline{\includegraphics[trim=180 85 245 70,clip, width=1.0\linewidth]{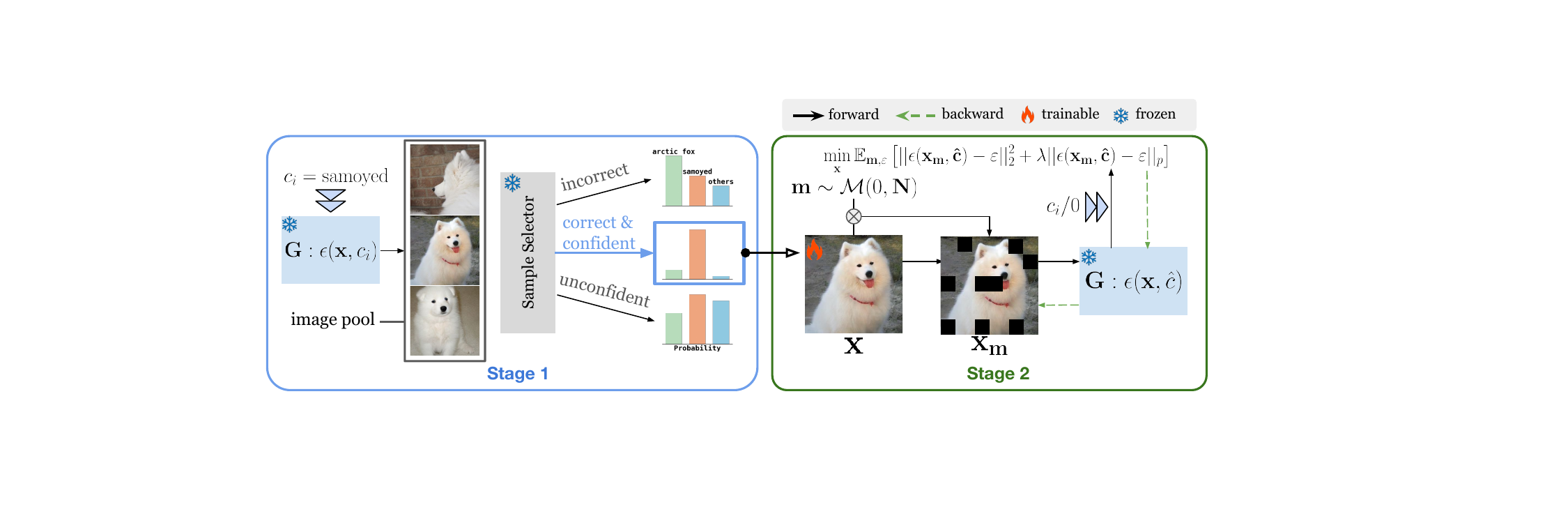}}
\vspace{-8pt}
\caption{Pipeline of our method when applied to the Masked Autoregressive model.}
\label{fig:mar_pipeline}
\end{center}
\vspace{-36pt}
\end{figure}

\cref{fig:mar_pipeline} shows the pipeline of how we utilize the MAR \cite{mar} backbone for our framework. The process differs from the DiT-based pipeline in two aspects: (1) Instead of perturbing the input latent using Gaussian noise w.r.t. random time steps, we perturb the input by randomly masking patches w.r.t. a maximum masking ratio; (2) The unconditional guidance is not available in MAR, thus we use a zero label embedding obtained by reformulating the $\texttt{Embedding()}$ layer as a linear layer. The first stage of sample selection is the same as that of \cref{fig:pipeline}.

We find that MAR exhibits stronger distillation performance than DiT, and is more efficient in both distillation time and GPU memory cost. We utilize the MAR-Base model, but observe that using larger versions such as MAR-Large and MAR-Huge does not lead to better performance. 

\begin{figure*}
\begin{center}
\centerline{\includegraphics[trim=100 85 150 55,clip, width=1.0\textwidth]{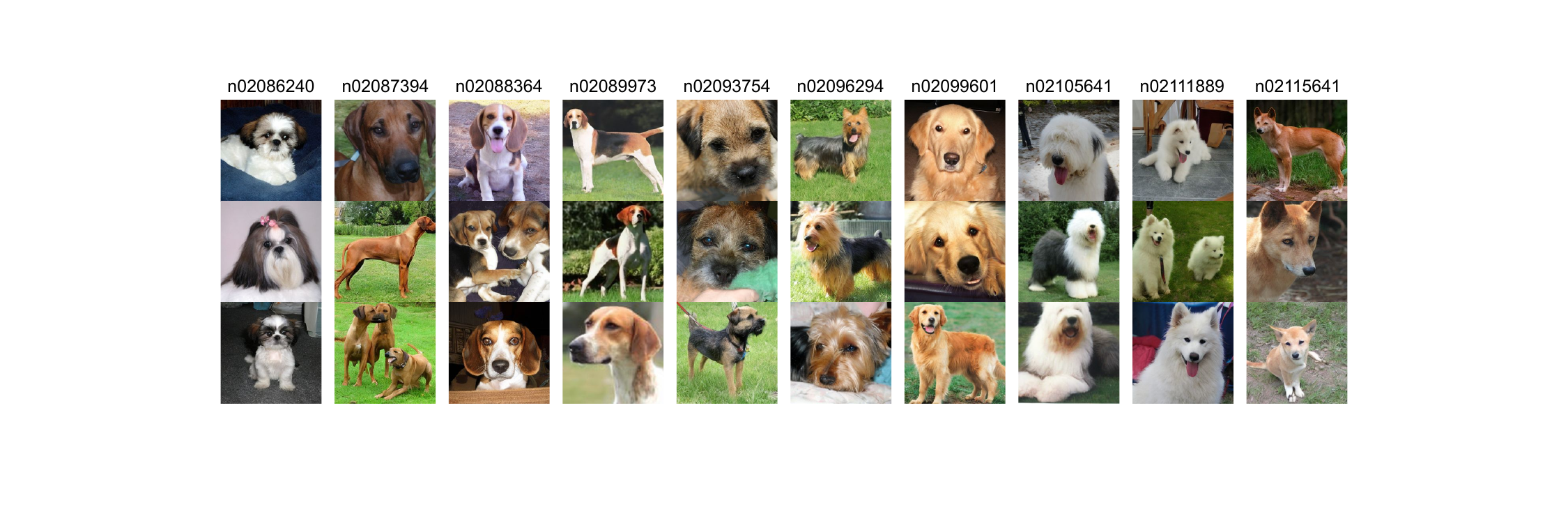}}
\vspace{-8pt}
\caption{More examples of our distilled images on ImageWoof.}
\label{fig:woof_supp}
\end{center}
\vspace{-18pt}
\end{figure*}

\section{More Baseline Comparisons}
\label{sec:exp_supplement}
\subsection{Quantitative Comparison with DD Methods}
\begin{table}[h]
    \centering
    \small
    \vspace{-8pt}
    \begin{minipage}{0.49\linewidth}
        \adjustbox{width=1.0\linewidth}{
            \begin{tabular}{c|cc}
            \toprule
                IPC & 10 & 50 \\
            \midrule
                G-VBSM & 31.4±0.5 & 51.8±0.4 \\
                EDC & \cellcolor{lightpurple}48.6±0.3 & 58.0±0.2 \\
                DiT-IGD &  45.5±0.5 & 59.8±0.3 \\
                \textbf{Ours} & 46.1±0.2 & \cellcolor{lightpurple}60.0±0.0 \\
            \bottomrule
            \end{tabular}
        }
        \vspace{-10pt}
        \caption{Baseline comparison on ImageNet-1K.}
        \label{tab:sota_comp}
    \end{minipage}
    \begin{minipage}{0.50\linewidth}
        \adjustbox{width=1.0\linewidth}{
        \begin{tabular}{c|cc}
        \toprule
        Woof  & IPC=10 & IPC=50 \\
        \midrule
            DiT-IGD & \cellcolor{lightpurple}67.7±0.3 & 81.0±0.7 \\
            \textbf{Ours} (w/o SS) & 65.0±0.7 & \cellcolor{lightpurple}84.5±0.6 \\
        \midrule
        Nette  & IPC=10 & IPC=50 \\
        \midrule
            DiT-IGD &  44.8±0.8 & 62.0±1.1 \\
            \textbf{Ours} (w/o SS) & \cellcolor{lightpurple}45.6±1.4 & \cellcolor{lightpurple}68.9±1.1 \\
        \bottomrule
        \end{tabular}
        }
        \vspace{-10pt}
        \caption{Comparison w/o SS.}
        \label{tab:no_ss}
    \end{minipage}
    \vspace{-12pt}
\end{table}

In \cref{tab:sota_comp}, we report performance on ImageNet-1K using ResNet-18. We adopt IGD's DiT version for fair comparison.
In \cref{tab:no_ss}, we further compare with IGD without using sample selection (SS), showing the standalone effectiveness of single-stage CaO$_2$.

In \cref{tab:cifar10}, we present CIFAR-10 results. DATM and PAD are strong trajectory matching methods but less efficient and scalable, representing a different paradigm from us.
\begin{table}[h]
    \centering
    \small
    \vspace{-8pt}
    \begin{minipage}{0.8\linewidth}
        \adjustbox{width=1.0\linewidth}{
            \begin{tabular}{c|ccccc}
        \toprule
            IPC & SRe$^2$L & \cellcolor{lightpurple}Ours & DATM & PAD \\
        \midrule
            10 & 29.3±0.5 & \cellcolor{lightpurple}39.0±1.5 & 66.8±0.2 & 76.1±0.3 \\
            50 & 45.0±0.7 & \cellcolor{lightpurple}64.0±0.9 & 67.4±0.3 & 77.0±0.5 \\
        \bottomrule
        \end{tabular}
        }
        \vspace{-10pt}
        \caption{Comparison on CIFAR10.}
        \label{tab:cifar10}
        \vspace{-12pt}
    \end{minipage}
    \vspace{-4pt}
\end{table}

\subsection{Discussion on More Related Works}
LD3M \cite{ld3m} is similar to GLaD but replaces the GAN backbone with a diffusion model, combining matching-based approaches (e.g., MTT) with objectives that align latents to real datasets. In contrast, our method is orthogonal, as we avoid dataset matching and instead focus on fully leveraging the diffusion model, leading to improved efficiency and scalability. YOCO \cite{yoco} and BiLP \cite{bilp} use sample selection as preprocessing for matching-based DD to improve efficiency and denoise source data, while our method acts as a post-processing step tailored to diffusion-based DD. We also tested their protocols (EL2N, LBPE) and observed up to a 3\% performance drop compared to our design.

\section{More Ablations}
\label{sec:ab_image}
\noindent\textbf{Hyperparameter Recommendations.}
Though we ablated on the various hyperparameters, only a few need to be tuned. It is recommended to always use $L_\infty$ with $\lambda=10$, and pool size of $2/4 \times$ IPC.

\noindent\textbf{Time Cost Comparison.}
We provide the quantitative results for \cref{fig:time_acc} here:
\begin{table}[h]
    \centering
    \adjustbox{width=0.7\linewidth}{
    \begin{tabular}{c|cccc}
    \toprule
    IPC & SRe$^{2}$L & DiT & Minimax & Ours \\ \midrule
    1 &  299 & 12 & 3967 & 99 \\
    10 & 2392 & 81 & 4036  & 960 \\
    50 & 11338 & 388 & 4343 & 3958 \\
    \bottomrule
    \end{tabular}
    }
    \vspace{-8pt}
    \caption{Time Cost (s) Comparison on ImageWoof.}
    \label{tab:time_fig1}
    \vspace{-8pt}
\end{table}

\noindent\textbf{Level of noise perturbation.} Beyond selection strategy and condition choice, we also investigate the impact of varying noise perturbation levels in the latent optimization process. Greater perturbation severity introduces noisier image input during latent optimization, thereby increasing denoising difficulty and accentuating key semantic features.
The degree of perturbation is determined by the maximum time step $\hat{T}$, where we randomly sample $t \sim [1, \hat{T}]$. A larger $\hat{T}$ increases the amount of noisier inputs during latent optimization.
Let $T$ represent the total number of time steps; the impact of noise level is detailed in \cref{tab:t_effect}.

\begin{table}[h]
\centering
\small
\adjustbox{width=0.8\linewidth}{
\begin{tabular}{ccc}
\toprule
$\hat{T}$   & ImageWoof IPC=10 & ImageNette IPC=10 \\ \midrule
$T$/12 & 42.7±0.8    & 61.9±1.6     \\
$T$/8  & \textbf{44.4±0.2}    & 61.9±1.6     \\
$T$/4  & 42.3±1.0    & 62.9±1.0     \\
$T$/2  & 42.3±1.6    & \textbf{63.5±0.8}     \\
$T$ & 42.7±0.7    & 62.3±0.7     \\ \bottomrule
\end{tabular}
}
\vspace{-8pt}
\caption{Effect of the noise perturbation level.}
\label{tab:t_effect}
\vspace{-4pt}
\end{table}

From the table, we observe that for challenging tasks like ImageWoof, a lower level of noise perturbation is more advantageous, while for easier tasks like ImageNette, a relatively higher noise level is beneficial. Additionally, an extremely low noise level yields sub-optimal performance, as does using all time steps. We speculate that this is because the latent optimization process requires a minimum noise level to improve image robustness. For harder tasks, optimization should be conservative to avoid shifting images toward the region of another class, while for easier tasks, a more aggressive approach enhances discriminative features.

\noindent\textbf{Effect of stage ordering.} We analyze the ordering the of the current stage designs. As shown in \cref{tab:order}, reversing the stages reduces performance and increases distillation time due to the additional latents requiring optimization.
\begin{table}[h]
    \centering
    \small
    \vspace{-8pt}
    \adjustbox{width=1.0\linewidth}{
    \begin{tabular}{c|cccc}
    \toprule
    Acc (\%) / Time (min)  &  \begin{tabular}[c]{@{}c@{}}Woof\\ IPC=10\end{tabular} &  \begin{tabular}[c]{@{}c@{}}Woof\\ IPC=50\end{tabular} & \begin{tabular}[c]{@{}c@{}}Nette\\ IPC=10\end{tabular} & \begin{tabular}[c]{@{}c@{}}Nette\\ IPC=50\end{tabular} \\
    \midrule
    \cellcolor{lightpurple}\textbf{CaO}$_\mathbf{2}$  & \cellcolor{lightpurple}45.6 / 15  & \cellcolor{lightpurple}68.9 / 64 & \cellcolor{lightpurple}65.0 / 15 & \cellcolor{lightpurple}84.5 / 64 \\ 
    Reverse  & 37.3 / 46 & 67.7 / 115 & 61.9 / 46 & 83.0 / 115 \\   
       \bottomrule
    \end{tabular}
    }
    \vspace{-10pt}
    \caption{Effect of stage ordering.}
    \label{tab:order}
    \vspace{-4pt}
\end{table}

\noindent\textbf{Superiority of using generated images.} We justify when generated synthetic images may be a better solution than randomly sampled real images. \cref{tab:real_gen} shows that diffusion-generated images perform better than carefully selected real ones, especially under lower IPC settings. A similar phenomenon is also observed on ImageNette.
\begin{table}[h]
    \centering
    \small
    \vspace{-8pt}
    \adjustbox{width=1.0\linewidth}{
    \begin{tabular}{c|ccc|ccc}
    \toprule
    Acc & \multicolumn{3}{c|}{IPC=1} & \multicolumn{3}{c}{IPC=10} \\
    (\%) & R18 & R50 & R101 & R18  & R50 & R101 \\ \midrule
    Real  & 13.1±0.8 & 13.8±0.6 & 14.4±1.2 & 39.1±0.9 & 36.9±0.5 & 31.8±0.9 \\
    \cellcolor{lightpurple}\textbf{Gen}  & \cellcolor{lightpurple}19.5±0.8 & \cellcolor{lightpurple}19.9±0.5 & \cellcolor{lightpurple}20.0±0.9 & \cellcolor{lightpurple}42.6±1.1 & \cellcolor{lightpurple}38.5±0.3 & \cellcolor{lightpurple}36.4±1.1 \\
    \bottomrule
    \end{tabular}
    }
    \vspace{-10pt}
    \caption{Comparison on ImageWoof (same selection settings).}
    \label{tab:real_gen}
    \vspace{-12pt}
\end{table}

\noindent\textbf{Comparison with classifier-guided models.} \cref{tab:cfg} compares the performance of using classifier-guided models with classifier-free counterparts. The reasons we do not use classifier-guided models are threefolds: (1) From the table, we see that guided-diffusion empirically provides limited discriminative information, performing similarly to its classifier-free counterpart. (2) They also require additional classifiers, increasing parameters and being slower than a simple ResNet. (3) Most diffusion models are trained with CFG, thus we focus on this family of models to be more generalizable. 
\begin{figure}[h]
\centering
    \adjustbox{width=0.5\linewidth}{
        \begin{tabular}{c|cc}
        \toprule
         &  IPC=10 &  IPC=50 \\
        \midrule
        CG & 42.6±0.7 & 67.2±0.6 \\ 
        CFG & 43.3±1.9 & 66.8±1.5 \\   
           \bottomrule
        \end{tabular}
    }
        \vspace{-8pt}
        \caption{Comparison of using classifier-guidance or not.}
        \label{tab:cfg}
        \vspace{-12pt}
\end{figure}

\section{Influence of different evaluation paradigms}
\label{sec:eval_comp}
We compare the popularly used hard-label \cite{minimax} and soft-label \cite{rded} evaluation metrics in \cref{tab:minimax_comp}, using distilled images from Minimax Diffusion as an example. From the table, we show that neither of the two approaches can always obtain better performance. 

\begin{table}[h]
    \vspace{-8pt}
    \centering
    \adjustbox{width=1.0\linewidth}{
        \begin{tabular}{c|ccc|ccc}
        \toprule
        Setting & \begin{tabular}[c]{@{}c@{}}\\ IPC=1\end{tabular} &  \begin{tabular}[c]{@{}c@{}}Woof\\ IPC=10\end{tabular} & \begin{tabular}[c]{@{}c@{}}\\ IPC=50\end{tabular} & \begin{tabular}[c]{@{}c@{}}\\ IPC=1\end{tabular} &  \begin{tabular}[c]{@{}c@{}}Nette\\ IPC=10\end{tabular} & \begin{tabular}[c]{@{}c@{}}\\ IPC=50\end{tabular} \\
        \midrule
        Hard-label \cite{minimax} & \textbf{19.9±0.2} & 36.2±0.2 & 57.6±0.9 & \textbf{31.8±0.6} & 54.9±0.1 & 74.2±1.3 \\
        Soft-label \cite{rded} & 18.2±1.1 & \textbf{40.1±1.0} & \textbf{67.0±1.8} & 22.6±1.2 & \textbf{61.4±0.7} & \textbf{83.9±0.2} \\
        \bottomrule
        \end{tabular}
    }
    \vspace{-8pt}
    \caption{Comparison on Minimax images using ResNet18.}
    \label{tab:minimax_comp}
    \vspace{-12pt}
\end{table}

\begin{figure*}[h]
\begin{center}
\centerline{\includegraphics[trim=100 85 150 55,clip, width=1.0\textwidth]{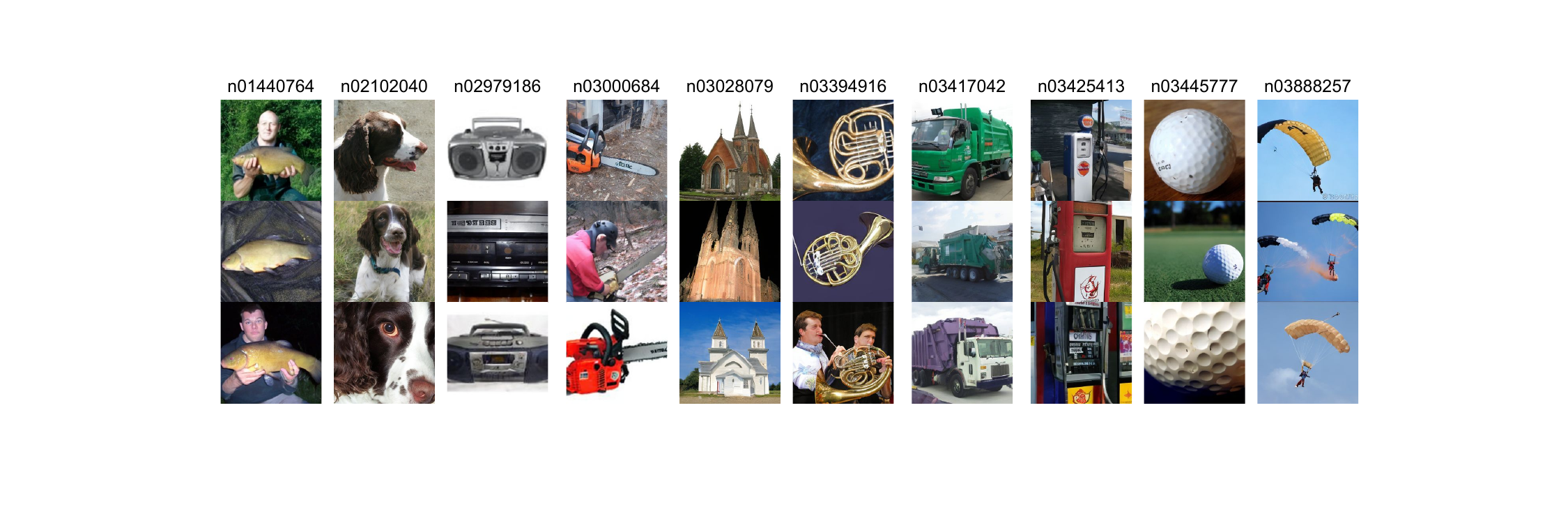}}
\vspace{-8pt}
\caption{Examples of our distilled images on ImageNette.}
\label{fig:nette_supp}
\end{center}
\vspace{-24pt}
\end{figure*}

We also observe other cases where using hard-labels outperform soft-labels:
\begin{itemize}
    \item For ResNet50 training on ImageNet-100 with Minimax images, using the ResNet18 model to generated soft-labels leads to only 1.0\% accuracy. This indicates that a good expert is critical for successful guidance.
    \item For ResNet101 training on ImageNet-1K (IPC=1) with our method, using hard-labels leads to $6.0\pm0.4$ accuracy while using soft-labels leads to $5.8\pm0.7$ accuracy. We induce that the prior knowledge from the expert may be insufficient when the IPC is low.
\end{itemize}

From the above results, we conclude that there is currently no unified evaluation paradigm that is being simultaneously effective, stable, and does not require external prior knowledge. Relevant works such as DD-Ranking \cite{ddranking} were developed, but yet (March 2025) does not support ImageNet-level datasets. Benchmarking and unifying the distilled datasets remains an open question and is of vital importance.

\section{Additional Visualizations}
\label{sec:visual}
We provide more visualization results here for a comprehensive analysis of our method.

\noindent\textbf{Distilled images of ImageWoof and ImageNette.} \cref{fig:woof_supp} and \ref{fig:nette_supp} show examples of distilled images under IPC=10 for ImageNette and ImageWoof. Three samples are shown for each category. From the distilled images, we see that our method effectively covers the class distribution and produces high-fidelity images. One thing we noticed is that although the classification performance on ImageNette is significantly higher than that of ImageWoof, the sample quality of both tasks is similar. The reason is straightforward: the categories in ImageNette are distinct, and therefore, easily distinguishable.
This observation indicates that the class composition of a task matters, suggesting that more attention should be paid to the tasks than to the individual classes during distillation, supporting the design of our approach.

\begin{figure}[h]
\begin{center}
\centerline{\includegraphics[trim=325 85 370 55,clip, width=1.0\linewidth]{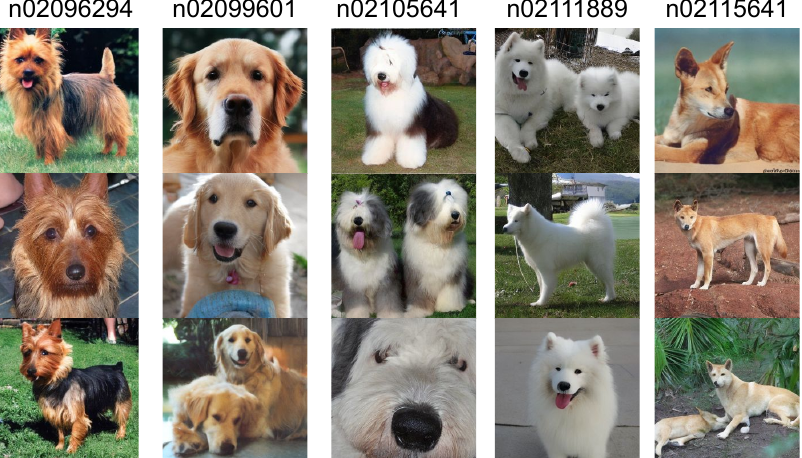}}
\vspace{-8pt}
\caption{Examples of our distilled images when using the Minimax Diffusion model as backbone.}
\label{fig:minimax_supp}
\end{center}
\vspace{-24pt}
\end{figure}

\noindent\textbf{Distilled images with Minimax Diffusion backbone.}
We further provide examples of the images generated via the Minimax backbone. \cref{fig:minimax_supp} shows examples of the distilled images in ImageWoof. Compared to the DiT backbone, the use of the Minimax Diffusion backbone further enhances the diversity of the distilled images. This phenomenon also suggests the extensibility of our proposed method, indicating its applicability as a plug-and-play module for existing and future work.

\begin{figure}[h]
\begin{center}
\centerline{\includegraphics[trim=325 85 370 55,clip, width=1.0\linewidth]{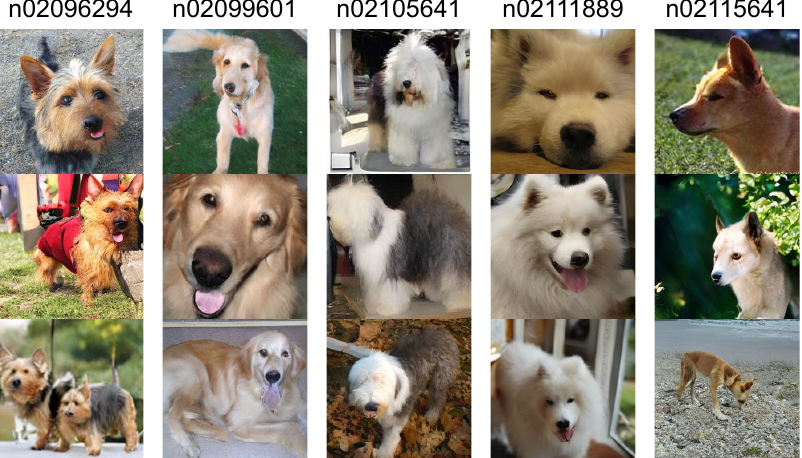}}
\vspace{-8pt}
\caption{Examples of our distilled images when using MAR as backbone.}
\label{fig:mar_supp}
\end{center}
\vspace{-24pt}
\end{figure}

\noindent\textbf{Distilled images with MAR backbone.} \cref{fig:mar_supp} presents example distilled images generated using MAR as model backbone. Interestingly, although MAR-distilled images achieve higher classification performance compared to those distilled with DiT, we observe that their image quality is generally lower. In fact, the images shown are those selected for their best visual quality.
We conjecture that the reason might be: although the overall image quality is low, the essential features related to the corresponding category are emphasized, while background and irrelevant features are de-emphasized. As a result, even if the images appear visually poor to human observers, they possess strong discriminative capabilities.

\noindent\textbf{More analysis on \cref{fig:examples}.} The optimization objective improves image-label consistency, refining \textit{category boundaries} to enhance class characteristics. \textit{Background} adjustments may occur because diffusion models, trained with a noise prediction objective, only fully denoise as $t \xrightarrow{} \infty$. Under limited NFE, generated latents remain partially denoised, and the changes likely result from removing residual noise.

\section{Limitations and Potential Improvements}
\label{sec:impact}
Although diffusion-based methods demonstrate strong performance, their applicability is constrained by the limited conditions these models can handle (e.g. DiTs can only deal with ImageNet classes). Employing text-to-image models such as Stable Diffusion can help mitigate this issue, but the large model size and absence of classification constraints may hinder practical application.Therefore, developing efficient and task-adaptive approaches based on text-to-image models might be a way to enable effective handling of arbitrary classes.
Moreover, the two inconsistencies we observe arise from the fundamental difference between generation and discrimination. Thus, developing a unified framework for both generation and classification may also significantly advance the field of diffusion-based dataset distillation.

\end{document}